\documentclass[10pt,twocolumn,letterpaper]{article}

\usepackage[utf8]{inputenc}

\usepackage[pagenumbers]{cvpr} 

\usepackage{tabularx}

\definecolor{cvprblue}{rgb}{0.21,0.49,0.74}
\usepackage[pagebackref,breaklinks,colorlinks,allcolors=cvprblue]{hyperref}
\newcommand{\email}[1]{\href{mailto:#1}{#1}}

\usepackage{xcolor}

\def\paperID{7} 

\def\confName{EarthVision}
\def\confYear{2025}

\title{Hybrid AI–Physical Modeling for Penetration Bias Correction in X-band InSAR DEMs: A Greenland Case Study}

\author{
    Islam Mansour$^{1,2}$, Georg Fischer$^{1}$, Ronny Hänsch$^{1}$, and Irena Hajnsek$^{1,2}$
    \\
    \tt\small $^1$Microwaves and Radar Institute, German Aerospace Center DLR, Germany 
    \\
    \tt\small $^2$Institute of Environmental Engineering, ETH Zurich, Switzerland\\
    {\tt\small 
        \tt\small \email{islam@imansour.net}, 
        \tt\small \email{georg.fischer@dlr.de}, 
        \tt\small \email{rww.haensch@gmail.com}, 
        \tt\small \email{irena.hajnsek@dlr.de}
    }
}

\begin{document}
\maketitle
\begin{abstract}
    Digital elevation models derived from Interferometric Synthetic Aperture Radar (InSAR) data over glacial and snow-covered regions often exhibit systematic elevation errors, commonly termed "penetration bias." We leverage existing physics-based models and propose an integrated correction framework that combines parametric physical modeling with machine learning. We evaluate the approach across three distinct training scenarios — each defined by a different set of acquisition parameters — to assess overall performance and the model's ability to generalize. Our experiments on Greenland's ice sheet using TanDEM-X data show that the proposed hybrid model corrections significantly reduce the mean and standard deviation of DEM errors compared to a purely physical modeling baseline. The hybrid framework also achieves significantly improved generalization than a pure ML approach when trained on data with limited diversity in acquisition parameters.\footnote{The source code is available at \url{https://github.com/IslamAlam/pydeepsar}}
\end{abstract}

\section{Introduction}

\begin{figure}[!htb]
        \centering
        \includegraphics[width=0.95\linewidth]{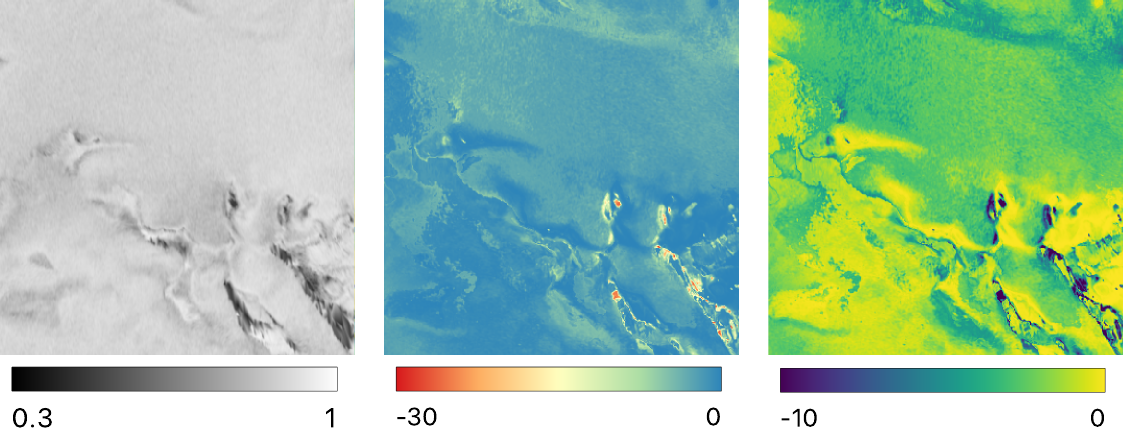} 
    \caption{Qualitative example from a representative region. 
    \emph{Left}: Interferometric coherence (range: 0.3--1). 
    \emph{Center}: One-way penetration depth ($d_{\mathrm{pen}}$, in meters) predicted by the MLP component of our hybrid framework, which then feeds into the Exponential profile. 
    \emph{Right}: Final penetration bias ($p_{\mathrm{bias}}$, in meters). 
    This hybrid approach leverages InSAR features (e.g., coherence) to predict $d_{\mathrm{pen}}$, which is subsequently used to estimate $p_{\mathrm{bias}}$ for correcting DEM elevations over ice and snow.}
    \label{fig:coherence_dpen_bias}
\end{figure}

Interferometric Synthetic Aperture Radar (InSAR) has enabled the generation of high-resolution Digital Elevation Models (DEMs) over large areas, such as the global Copernicus DEM derived from TanDEM-X \cite{noauthor_copernicus_2022}. These DEMs are crucial for many applications, including cryospheric studies, where they enable monitoring of glacier mass balance and ice sheet dynamics \cite{abdullahi_estimating_2019}.

One of the main challenges in generating InSAR DEMs is the penetration of the radar signal into dry snow and ice. This causes the measured elevation to lie several meters below the actual surface, leading to the so-called penetration bias~\cite{dall_topography_2001, rignot_penetration_2001, wessel_new_2016, fan_new_2022}. For example, a recent study by Fan et al.~\cite{fan_new_2022} compared DEMs derived from ICESat-2 (a laser altimeter sensor) and TanDEM-X (an X-band radar sensor) data over Greenland and found that the elevation bias varies significantly with terrain elevation. In their analysis, for elevations above 2000~m the median difference (MED) between the DEMs is approximately -3.76~m (with an RMSE of 4.51~m), whereas for elevations below 2000~m the MED is around -2.32~m (with an RMSE of 7.00~m), yielding an overall MED of -2.75~m and RMSE of 6.58~m. This pronounced bias — primarily due to X-band signal penetration into the ice — illustrates both the inherent challenge of penetration bias and the influence of topography, thereby motivating the need for advanced correction techniques. 

There is potential to estimate the penetration bias directly from the InSAR data, because the signal penetration depends on the snow and ice properties, which, in turn, influence the backscatter and coherence measured by the InSAR system. The key relationship is between the vertical scattering profile, which describes how the backscattered signals are distributed within the snow and ice, and the interferometric coherence, which measures the correlation between the InSAR acquisition pair. Typically, a shallow vertical scattering profile leads to high coherence, whereas deep penetration causes low coherence. This relationship depends on the InSAR acquisition geometry that determines the interferometric height sensitivity, which is described by the Height of Ambiguity (HoA) parameter.

Physical models, such as the Uniform Volume (UV) model (which assumes a uniform signal extinction in snow and ice, i.e., an exponential vertical profile), are used to estimate the penetration depth under idealized conditions~\cite{weber_hoen_penetration_2000, dall_insar_2007, fischer_modeling_2020}. One key advantage of these physics-based methods is that they do not require any additional training or reference data, unlike data-driven approaches that depend on high-quality reference data. However, these models can over- or underestimate the bias when real scattering scenarios deviate from their simplifications. In contrast, purely data-driven approaches, such as machine learning (ML) models, can capture local variations more accurately and flexibly but may lack the robustness and interpretability of physics-based methods \cite{abdullahi_estimating_2019, campos_potential_2024}.

Building upon our initial hybrid framework~\cite{mansour_correction_2024}, we present an expanded analysis that synergistically combines physical models with ML approaches. While our previous work focused solely on the Exponential profile, here we systematically evaluate two parametric physical models (Exponential and Weibull profiles) to comprehensively understand their capabilities and limitations in modeling vertical scattering profiles. We train a Multi-Layer Perceptron (MLP) to predict the parameters of these physical models from the data, as illustrated in Figure~\ref{fig:coherence_dpen_bias}. To rigorously assess generalization capabilities, we introduce three distinct HoA scenarios: (1) training with all available HoA scenes, (2) an \emph{Interpolation scenario} where we exclude mid-range HoA values to test interpolation capabilities, and (3) an \emph{Extrapolation scenario} where we exclude higher HoA values to evaluate extrapolation performance. This systematic evaluation provides crucial insights into model robustness and generalization capabilities across variable acquisition geometries as expected for large scale InSAR applications.

\section{Related Work}
\label{sec:related}

\subsection{Physical Modeling Approaches}
Accurate electromagnetic modeling of the vertical scattering profile in snow and ice is a critical prerequisite for correcting the penetration bias in InSAR-derived DEMs. The interaction of microwaves with natural snow cover is complex because the snow comprises heterogeneous mixtures of ice grains and air that also depend on environmental conditions such as temperature and metamorphism processes. Early studies provide insights into the dielectric properties and scattering mechanisms in natural snow \cite{matzler_applications_1987}. 

The Uniform Volume (UV) is introduced as a simplified model for the scattering medium \cite{weber_hoen_penetration_2000}. It treats the snow and ice volume as an infinitely deep, homogeneous medium with a constant signal extinction coefficient, leading to an exponential vertical scattering profile. Under these assumptions, the penetration bias can be estimated directly from the InSAR data using the UV model~\cite{dall_insar_2007,sharma_estimation_2013}.
Another more flexible solution is based on the Weibull profile \cite{fischer_modeling_2020} capturing changes in signal extinction due to increasing grain size and density with depth \cite{matzler_applications_1987}. It is able to compensate for the underestimation of the UV model, but it is also prone to overestimating the surface location \cite{fischer_modeling_2020}.

\subsection{Data-Driven Approaches}
Purely data-driven methods bypass explicit physical modeling by learning the mapping between observable SAR features (e.g., interferometric coherence and backscatter intensity) and the target variable (e.g., the penetration bias, snow-facies) directly from training data.

A key study proposes a pixel-based regression approach for estimating X-band InSAR elevation bias over Greenland \cite{abdullahi_estimating_2019}.
Their method achieves a coefficient of determination ($R^2$) of 0.68 and an RMSE of 0.68~m. While computationally efficient, this approach is limited by its linearity and lack of explicit physical constraints.

Other recent work explores deep learning-based methods, for example, using a deep unsupervised learning approach for the classification of snow facies \cite{becker_campos_unsupervised_2024, campos_potential_2024}. In this case, the snow facies are segmented and used for the penetration bias estimation. 

Although these methods can capture complex empirical relationships, they generally require extensive, high-quality training datasets — often derived from sources such as CryoSat-2 or ICESat-2 — that are very limited. Furthermore, purely data-driven approaches often struggle to generalize to conditions outside the training domain. The absence of explicit physical constraints also limits their interpretability.

\subsection{Hybrid Approaches}
Hybrid inversion methods combine model-based and data-driven approaches to leverage their advantages~\cite{Kurz:2022gtq}. This fusion, called hybrid modeling, combines physical model interpretability and rigor with ML's flexibility and adaptiveness. Physics-informed ML embeds physical constraints into learning algorithms, ensuring models fit data and adhere to physical laws~\cite{karniadakis_physics-informed_2021}. This approach has proven effective in complex applications like fluid dynamics and climate modeling~\cite{krasnopolsky_complex_2006, weisz_hybrid_2020, gibson_training_2021}. It holds a potential for inverting geo- and biophysical parameters from SAR data, overcoming limitations of purely physical or data-driven methods.

Recent work on the parameterization of the vertical scattering profile for forest height estimation has demonstrated that hybrid approaches can significantly improve the estimation of geophysical parameters~\cite{mansour_hybrid_2025}. However, despite its potential, specific work on physics-informed machine learning for InSAR parameter retrieval is lacking in current literature. Our work builds on these ideas by embedding the Exponential and Weibull profiles into a hybrid architecture. We compare the performance of these hybrid models with pure ML models for different HoA training scenarios, showing that incorporating physical constraints leads to improved accuracy, stability, and robustness — mainly when high-quality training data are limited.
\section{Methodology}
\label{sec:method}

\subsection{Problem Formulation}
Let \(h_{\text{InSAR}}\) be the InSAR-derived elevation and \(h_{\text{ref}}\) the reference “true” elevation at a given location. We define the penetration bias as
\begin{equation}
    \label{eq:bias}
    p_{\text{bias}} \;=\; h_{\text{InSAR}} - h_{\text{ref}},
\end{equation}
which must be estimated and removed to improve DEM accuracy in glaciated regions.

\subsection{Physical Models (Exponential, Weibull)}
\label{sec:phys_models}
The estimation of penetration bias is grounded in physical models that link the InSAR observation space to the vertical scattering profile in snow and ice — governed by geophysical parameters such as density, structure, and grain size.

In single-polarization InSAR, the primary observable is the complex coherence, $\widetilde{\gamma}$, which represents the cross-correlation between two acquisitions, $s_1$  and $s_2$ and is defined as~\cite{cloude_polarimetric_1998, rosen_synthetic_2000, cloude_polarisation_2009}: 
\begin{equation}
    \widetilde{\gamma}(\kappa_z)
    \;=\;
    \frac{\langle s_1 \, s_2^* \rangle}{\sqrt{\langle s_1 \, s_1^* \rangle\,\langle s_2 \, s_2^* \rangle}},
\end{equation}
where \(\kappa_z\) is the vertical wavenumber, \(\langle \cdot \rangle\) denotes an ensemble (or spatial) average, and \(^{*}\) indicates the complex conjugate. The observed coherence is factorized into several decorrelation terms, and can be expressed as~\cite{askne_c-band_1997, zebker_decorrelation_1992, hagberg_repeat-pass_1995, bamler_synthetic_1998}:
\begin{equation}
    \widetilde{\gamma}(\kappa_z)
    \;=\;
    \gamma_{Tmp}\,\gamma_{Rg}\,\gamma_{Sys}\,\widetilde{\gamma}_{\mathrm{Vol}}(\kappa_z),
\end{equation}
where \(\gamma_{Tmp}\) accounts for temporal decorrelation, \(\gamma_{Rg}\) represents range spectral decorrelation, \(\gamma_{Sys}\) captures system-related decorrelation, and \(\widetilde{\gamma}_{\mathrm{Vol}}(\kappa_z)\) corresponds to the volume decorrelation due to subsurface scattering. For TanDEM-X single-baseline InSAR, temporal decorrelation does not occur (\(\gamma_{Tmp}=1\)) and the other decorrelation factors are known~\cite{krieger_tandem-x_2007, gatelli_wavenumber_1994, martone_quantization_2015}.

To estimate the penetration bias \(p_{\text{bias}}\), we focus on the volume decorrelation \(\widetilde{\gamma}_{\mathrm{Vol}}(\kappa_z)\), which is directly linked to the vertical scattering profile \(f(z)\) describing how backscattered power varies with depth \(z\). For a semi-infinite volume (\(z \leq 0\)), the volume decorrelation is given by~\cite{sharma_estimation_2013}:
\begin{equation}
  \widetilde{\gamma}_{\mathrm{Vol}}(\kappa_z)
  \;=\;
  e^{j\,\kappa_z\,z_0}
  \,\frac{\displaystyle \int_{-\infty}^0 f(z)\,\exp\!\bigl(j\,\kappa_{z}\,z\bigr)\,dz}{\displaystyle \int_{-\infty}^0 f(z)\,dz},
  \label{eq:vol_coh_model}
\end{equation}
where \(z_0\) is the true surface height. The phase of \(\widetilde{\gamma}_{\mathrm{Vol}}(\kappa_z)\), denoted as \(\phi_{\mathrm{Vol}} = \arg\!\Bigl(\widetilde{\gamma}_{\mathrm{Vol}}(\kappa_z)\Bigr)\), corresponds to the interferometric phase center, located in the subsurface. The penetration bias \(p_{\text{bias}}\) is then calculated as
\begin{equation}
    p_{\text{bias}}
    \;=\;
    \frac{\phi_{\mathrm{Vol}}}{\kappa_z},
    \label{eq:phase2bias}
\end{equation}
assuming \(z_0 = 0\) for simplicity. In practice, a known offset may be incorporated if \(z_0\) differs from a reference DEM. 

The detailed definitions of the vertical wavenumber \(\kappa_z\) and the HoA — which quantifies the interferometric phase-to-height sensitivity — are provided in the Supplementary Material (see Section~\ref{sec:hoa_supplement}, Eqs.~\ref{eq:kappa_z_def} and \ref{eq:HoA_def}). Briefly, these parameters relate the observed phase to height and are used to compute the elevation from the phase center.

\begin{figure}[!tb]
    \centering
    \begin{minipage}[b]{0.48\linewidth}
        \centering
        \includegraphics[width=\linewidth]{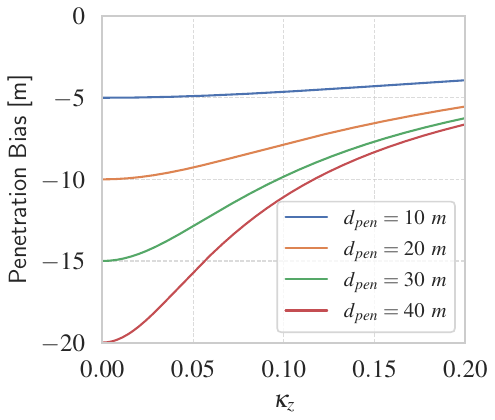}
        \caption*{\textbf{(a)} Exponential Profile}
    \end{minipage}
    \hfill
    \begin{minipage}[b]{0.48\linewidth}
        \centering
        \includegraphics[width=\linewidth]{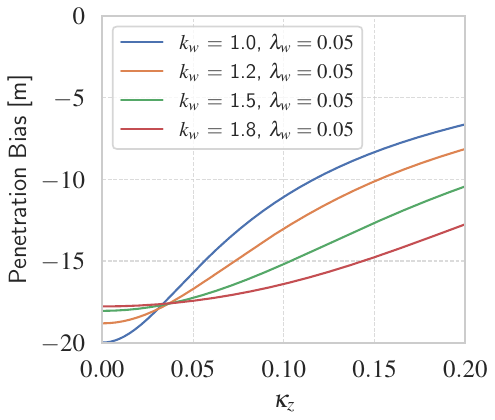}
        \caption*{\textbf{(b)} Weibull Profile}
    \end{minipage}
    \vspace{4pt}
    \caption{Penetration bias simulations for two scattering profiles, assuming a semi-infinite volume. 
    \textbf{(a)}~Exponential Profile with varying penetration depth $d_{\mathrm{pen}}$. 
    \textbf{(b)}~Weibull Profile for different shape parameters $k_w$ and fixed scale $\lambda_w=0.05$. 
    In both, the curves illustrate the penetration bias variation as a function of vertical wavenumber $\kappa_{z}$.}
    \label{fig:exponential_and_weibull}
\end{figure}

The physical models assume a specific form for the vertical scattering profile \(f(z)\). The UV model leads to an \emph{Exponential profile}, which is described as~\cite{weber_hoen_penetration_2000}:
\begin{equation}
    f_{\mathrm{exp}}(z)
    \;=\;
    \sigma_v^0\,\exp\!\Bigl(\frac{2z}{d_{\mathrm{pen}}}\Bigr),
    \label{eq:expprofile}
\end{equation}
where \(d_{\mathrm{pen}}\) is the one-way penetration depth and \(\sigma_v^0\) is a nominal scattering coefficient. Figure~\ref{fig:exponential_and_weibull} shows how the resulting penetration bias varies with different \(d_{\mathrm{pen}}\) values.

Under uniform-volume assumptions, one can derive a closed-form solution for \(\widetilde{\gamma}_{\mathrm{Vol}}(\kappa_z)\), making it possible to estimate the penetration bias \(p_{\mathrm{bias}}\) independently of \(d_{\mathrm{pen}}\). In this case, the volume decorrelation phase is given by \cite{dall_insar_2007}:
\begin{equation}
    \phi_{\mathrm{Vol}} \;=\;
    \tan^{-1}\!\Bigl(\sqrt{\tfrac{1}{|\gamma_{\mathrm{Vol}}|^2} - 1}\Bigr).
    \label{eq:uv_phase}
\end{equation}
We leverage this solution for computing \(p_{\mathrm{bias}}\) in our direct \emph{physical-model} estimation and use Eq.~\eqref{eq:vol_coh_model} with Eq.~\eqref{eq:expprofile} in the hybrid approach.

In contrast, the \emph{Weibull profile} is given by~\cite{fischer_modeling_2020}:
\begin{equation}
    f_{\mathrm{wb}}(z)
    \;=\;
    \lambda_w\,k_w\,(\lambda_w\,z)^{\,k_w - 1}\,\exp\!\Bigl(-(\lambda_w\,z)^{\,k_w}\Bigr),
    \label{eq:weibull}
\end{equation}
where \(\lambda_w\) (scale) and \(k_w\) (shape) allow for a more flexible representation of scattering behavior. Figure~\ref{fig:exponential_and_weibull} shows how different \(k_w\) values affect the penetration bias. However, estimating its two parameters, \(\lambda_w\) and \(k_w\), can be challenging. To ensure physically plausible solutions, we constrain \(\lambda_w\) to [0.01, 0.6] and \(k_w\) to [0.8, 1.5]. These parameter ranges provide vertical scattering profiles and associated penetration depths matching empirical data \cite{fischer_modeling_2019, fischer_modeling_2020}.

These models can be integrated into Eq.~\ref{eq:vol_coh_model} to compute the volume decorrelation \(\widetilde{\gamma}_{\mathrm{Vol}}(\kappa_z)\) and subsequently estimate the penetration bias \(p_{\text{bias}}\).

\begin{figure*}[!tb] 
    \centering
    \includegraphics[width=1\linewidth]{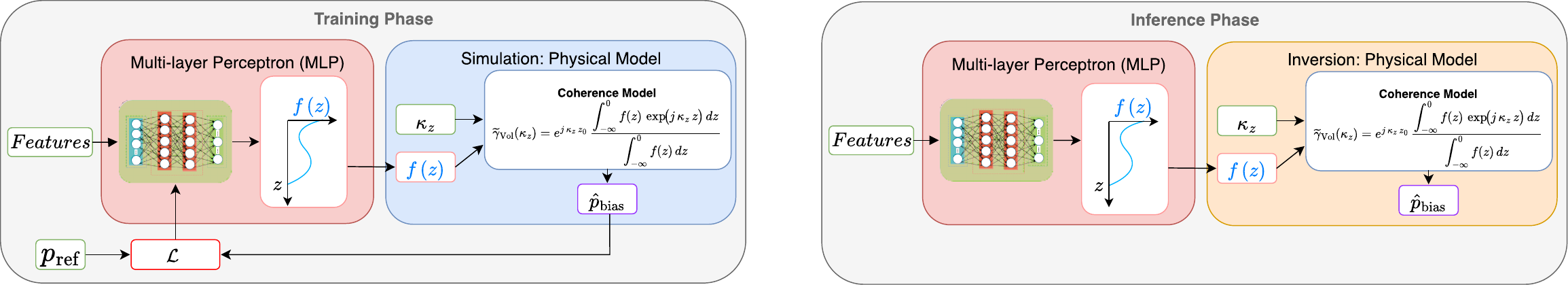}
    \caption{Overview of our hybrid modeling pipeline. An MLP predicts scattering profile parameters that feed into the physical model for computing the estimated bias \(\hat{p}_{\mathrm{bias}}\). We use MSE loss against a reference bias (e.g., LiDAR).}
    \label{fig:hybrid_model}
\end{figure*}

\subsection{Hybrid AI--Physical Model}
\label{sec:hybrid-model}
Our proposed hybrid framework (illustrated in Figure~\ref{fig:hybrid_model}) synergistically combines physical modeling with machine learning to estimate \(p_{\text{bias}}\) (Eq.~\ref{eq:bias}). Rather than relying solely on a fixed vertical scattering profile~$f(z)$ derived from idealized assumptions, we allow the model to learn the parameters that define \(f(z)\) from data. In practice, we train a MLP to predict the scattering profile parameters (e.g., \(\hat{d}_{\text{pen}}\) for the Exponential profile or \((\hat{\lambda}_w,\hat{k}_w)\) for the Weibull profile) from input features such as volumetric decorrelation, incidence angle, vertical wavenumber, interferometric phase, and backscatter. Given the low spatial variability in these homogeneous glaciated regions — and since even simple linear regression can estimate penetration bias with moderate accuracy — a straightforward MLP suffices, removing the need for more complex deep learning architectures \cite{abdullahi_estimating_2019}. This approach enables us to derive a unique solution that satisfies the physical constraints while adapting to real-world variability. The predicted parameters are then used in the volume decorrelation (Eq.~\eqref{eq:vol_coh_model}) to compute the estimated penetration bias \(\hat{p}_{\text{bias}}\). We train the hybrid model by minimizing the mean-squared-error (MSE) loss:

\begin{equation}
  \mathcal{L}_{\text{MSE}} 
  = \frac{1}{N} \sum_{n=1}^{N} \Bigl(\hat{p}_{\mathrm{bias},n} - p_{\mathrm{ref},n}\Bigr)^2,
    \label{eq:mse_loss}
\end{equation}
where \(p_{\mathrm{ref},n}\) denotes the reference bias for the \(n\)th sample and \(\hat{p}_{\mathrm{bias},n}\) represents the estimated penetration bias computed as in Eq.~\ref{eq:phase2bias}.

\section{Experiments}
\label{sec:experiments}

\subsection{Dataset}

\subsubsection{TanDEM-X InSAR Data}
\begin{figure}[!htb]
    \centering
    \includegraphics[width=0.95\linewidth]{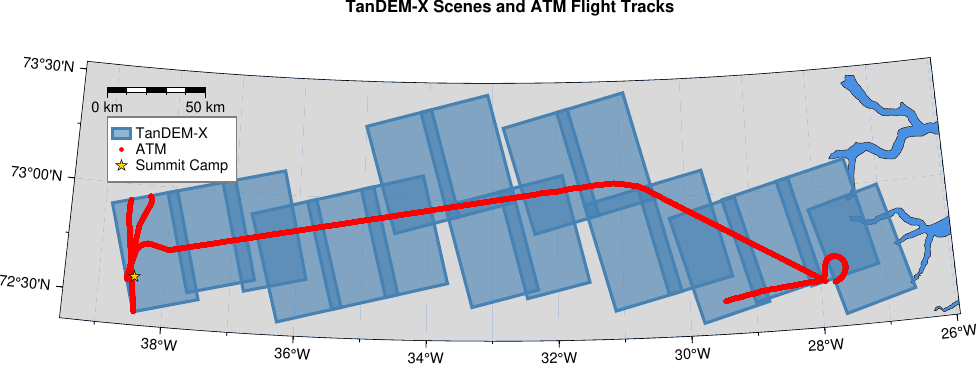}
    \caption{Overview of TanDEM-X scenes (blue) and ATM flight tracks (red) over the study area in Greenland. The Summit Camp location is marked with a yellow star.}
    \label{fig:tdx_atm_tracks}
\end{figure}

This study utilizes single-baseline InSAR imagery from the TanDEM-X mission to generate InSAR DEMs over Greenland. A total of \textbf{18} TanDEM-X CoSSC acquisitions are selected, covering a transect from the ice sheet summit to the East Coast, ensuring temporal and spatial alignment with NASA IceBridge data. Figure~\ref{fig:tdx_atm_tracks} provides an overview of the TanDEM-X scene footprints (in blue) and the corresponding NASA IceBridge ATM flight tracks (in red), acquired between January and May 2017 during the winter season.

\begin{figure*}[!htb]
\centering
\captionsetup{justification=centering}
\begin{subfigure}[b]{0.45\textwidth}
    \includegraphics[width=\linewidth]{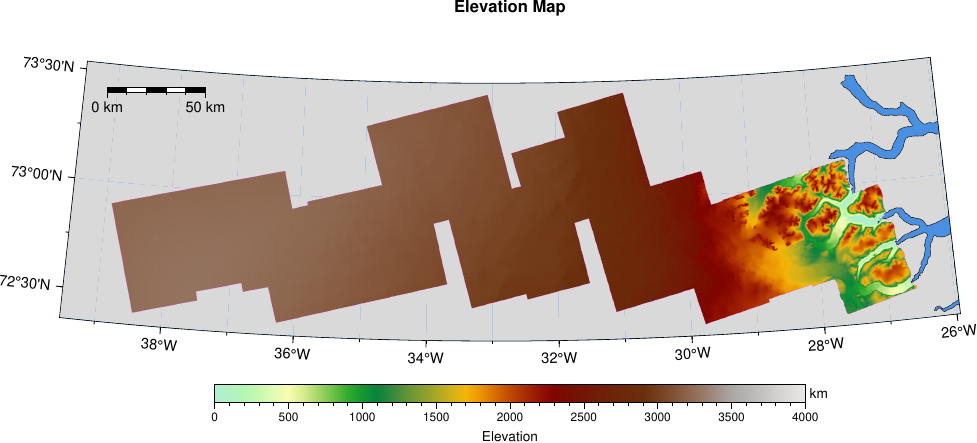}
    \caption{Elevation (DEM) from TanDEM-X}
\end{subfigure}
\hfill
\begin{subfigure}[b]{0.45\textwidth}
    \includegraphics[width=\linewidth]{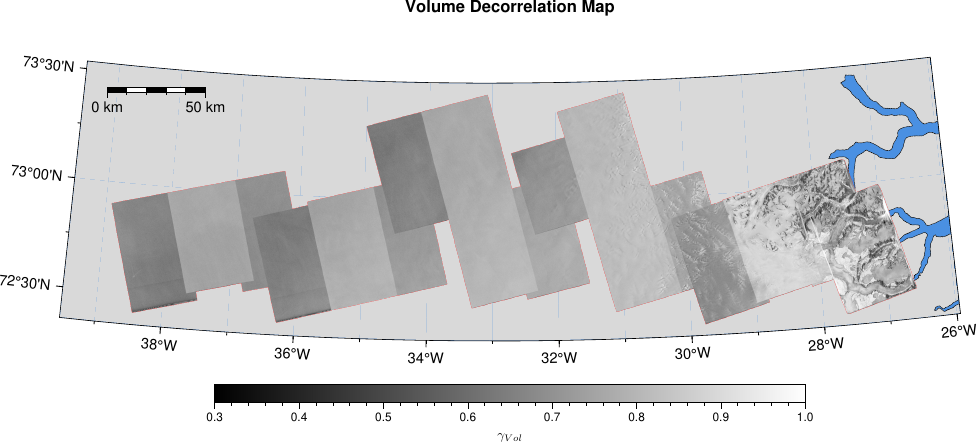}
    \caption{Volume Decorrelation}
\end{subfigure}

\vspace{1em}

\begin{subfigure}[b]{0.45\textwidth}
    \includegraphics[width=\linewidth]{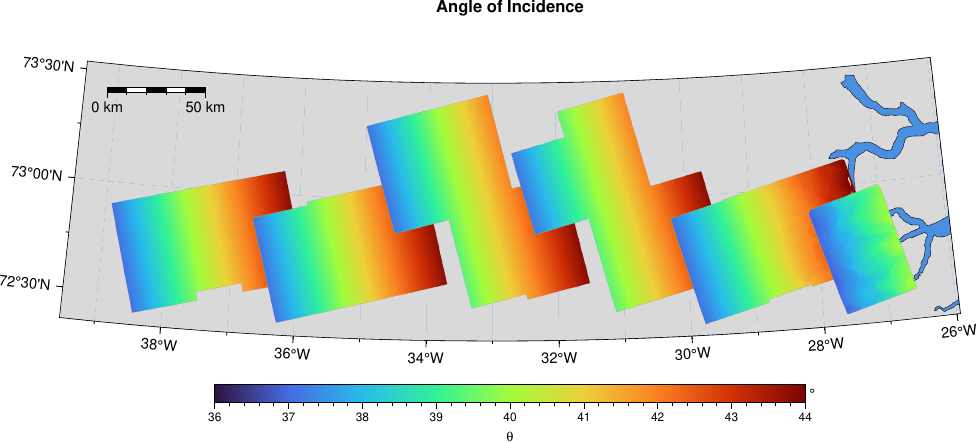}
    \caption{Incidence Angle}
\end{subfigure}
\hfill
\begin{subfigure}[b]{0.45\textwidth}
    \includegraphics[width=\linewidth]{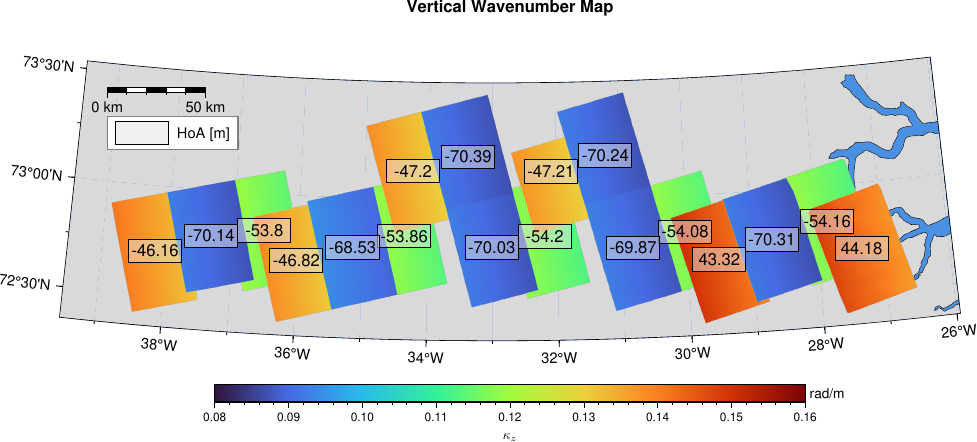}
    \caption{Vertical Wavenumber and HoA}
\end{subfigure}

\caption{Overview of the mosaicked dataset used in our study. Each panel shows a different attribute for the 2017 TanDEM-X acquisitions over Greenland.}
\label{fig:dataset_mosaic}
\end{figure*}

Post-processing steps include the derivation of InSAR elevation, coherence, backscatter, incidence angle, and vertical wavenumber. A mosaic overview of these acquisitions is shown in Figure~\ref{fig:dataset_mosaic}.

\subsubsection{NASA IceBridge ATM LiDAR Data}
\label{sec:atm}
For training and validation of the penetration bias correction model, we use the high-resolution surface elevation data from NASA's IceBridge Airborne Topographic Mapper (ATM)~\cite{alexandrov_icebridge_2018}. The ATM LiDAR dataset provides precise elevation measurements with a vertical uncertainty of less than 1\,m over flat ice surfaces. To ensure consistency with the InSAR-derived DEMs, the original 1\,m-resolution ATM dataset is resampled to the same grid as the InSAR products.

\begin{figure}[!htb]
    \centering
    \includegraphics[width=0.95\linewidth]{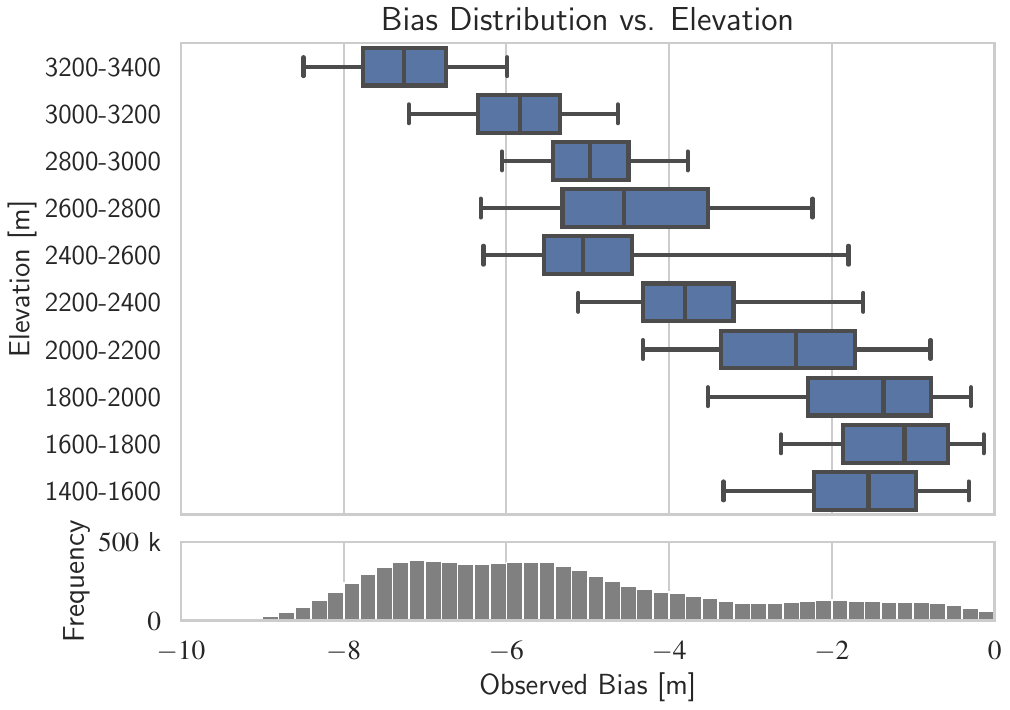} 
    \caption{Bias distribution across elevation bins, computed using ATM LiDAR as reference. The penetration bias increases at higher elevations due to smaller scattering losses from less melt-refreeze features in the subsurface.}
    \label{fig:bias_distribution}
\end{figure}

Leveraging these reference measurements, we also examine how the observed penetration bias varies with surface elevation. As shown in Figure~\ref{fig:bias_distribution}, higher elevations exhibit greater bias, as expected \cite{abdullahi_estimating_2019, fan_new_2022}, due to more dry snow with less melt-refreeze features in the subsurface allowing deeper X-band signal penetration.

\subsection{HoA-Based Training Scenarios}
\label{sec:hoa_scenarios}

To investigate how each method (hybrid or pure ML) generalizes to different HoA conditions, we define three training scenarios that include or exclude specific HoA ranges. We remove the scenes whose HoA values fall in the specified range (see Figure~\ref{fig:dataset_mosaic}\,(d) for an overview of the HoA distribution). The \emph{All} scenario uses every scene, \emph{Interpolation} excludes HoA in [50,\,60]\,m, and \emph{Extrapolation} excludes HoA above 70\,m.

We split the remaining data into training (60\%) and testing (40\%) sets, ensuring a broad distribution of ice sheet conditions. This approach reveals how well each model handles cases requiring interpolation (HoA gaps within the trained range) or extrapolation (HoA values beyond the trained range). In subsequent sections, we compare each scenario’s performance to assess model robustness and generalization.
For a more detailed view of the specific scenes included or excluded in each scenario, see Figure~\ref{fig:hoa_scenarios_supplement} in the Supplementary Material.

\begin{figure}[!htb]
    \centering
    \includegraphics[width=0.95\linewidth]{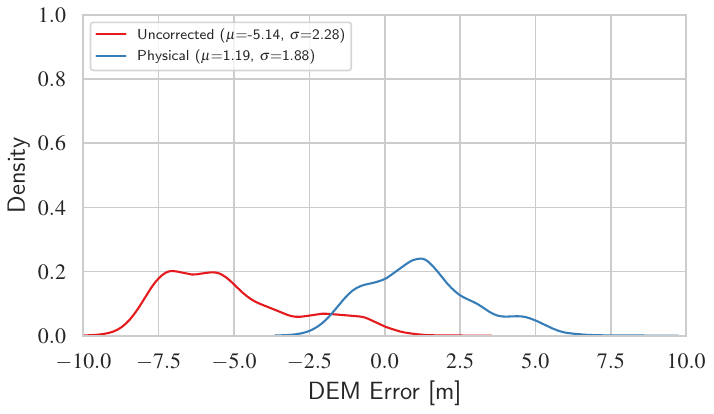}
    \caption{Error distributions for the uncorrected DEM (red) versus the physically corrected DEM (blue) using the UV direct solution of Eq.~\ref{eq:uv_phase}.}
    \label{fig:dem_correction_uv}
\end{figure}

\subsection{Metrics}
We evaluate the prediction accuracy using the mean error (ME), mean absolute error (MAE), mean absolute percentage error (MAPE), root mean square error (RMSE), and the coefficient of determination ($R^2$). These metrics quantify the estimator's bias, overall error magnitude, relative percentage error, and goodness of fit, respectively. In addition, we characterize the error distribution of the DEMs by computing the mean error (\(\mu\)) and the standard deviation (\(\sigma\)).
For the mathematical formulations of these metrics, please refer to the Supplementary Material (Section~\ref{sec:metrics_supplement}).

\subsection{Pure ML Approach (MLP)}
\label{sec:ml_model}
As an alternative to our hybrid framework, we implement a purely data-driven MLP model as a benchmark. This model directly regresses \(\hat{p}_{\text{bias}}\) from the input features, without incorporating explicit physical modeling of scattering physics.  We train the MLP by minimizing the same MSE loss defined in Eq.~\ref{eq:mse_loss}, where \(\hat{p}_{\mathrm{bias}}\) denotes the bias predicted by the MLP.


\begin{figure*}[!htb]
    \centering
    \begin{minipage}{0.30\textwidth}
        \centering
        \includegraphics[width=\linewidth]{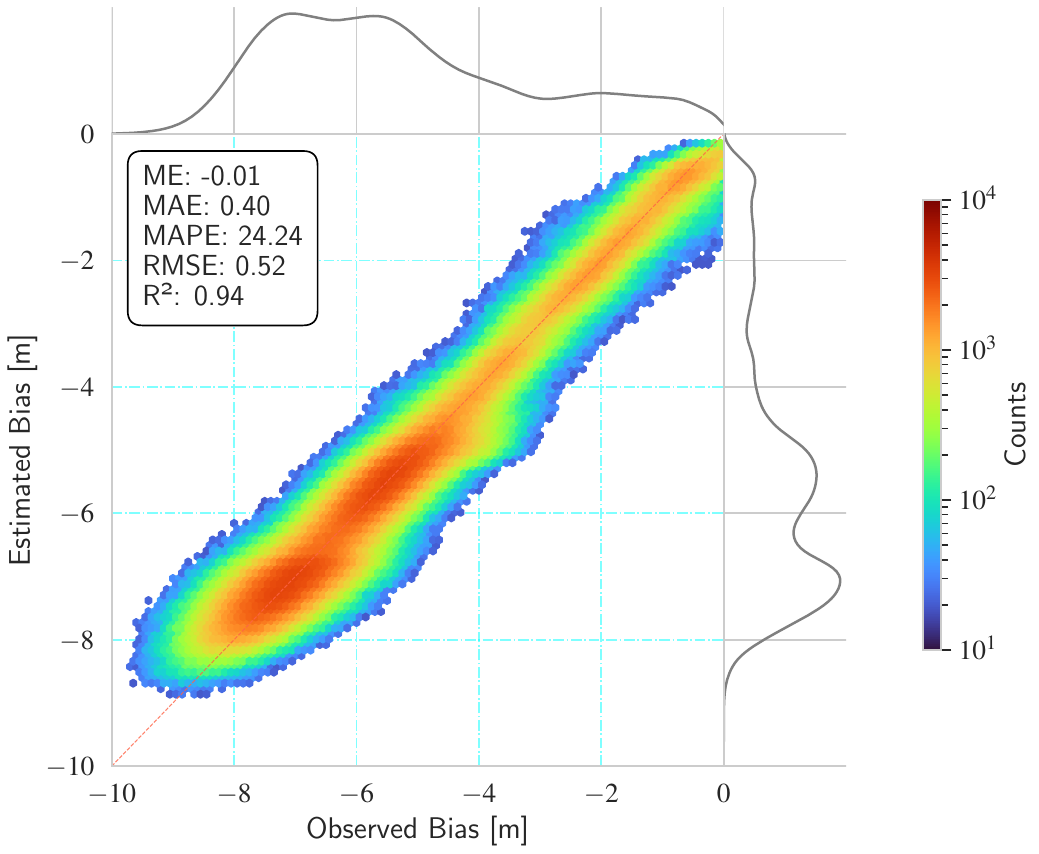}
    \end{minipage}
    \begin{minipage}{0.30\textwidth}
        \centering
        \includegraphics[width=\linewidth]{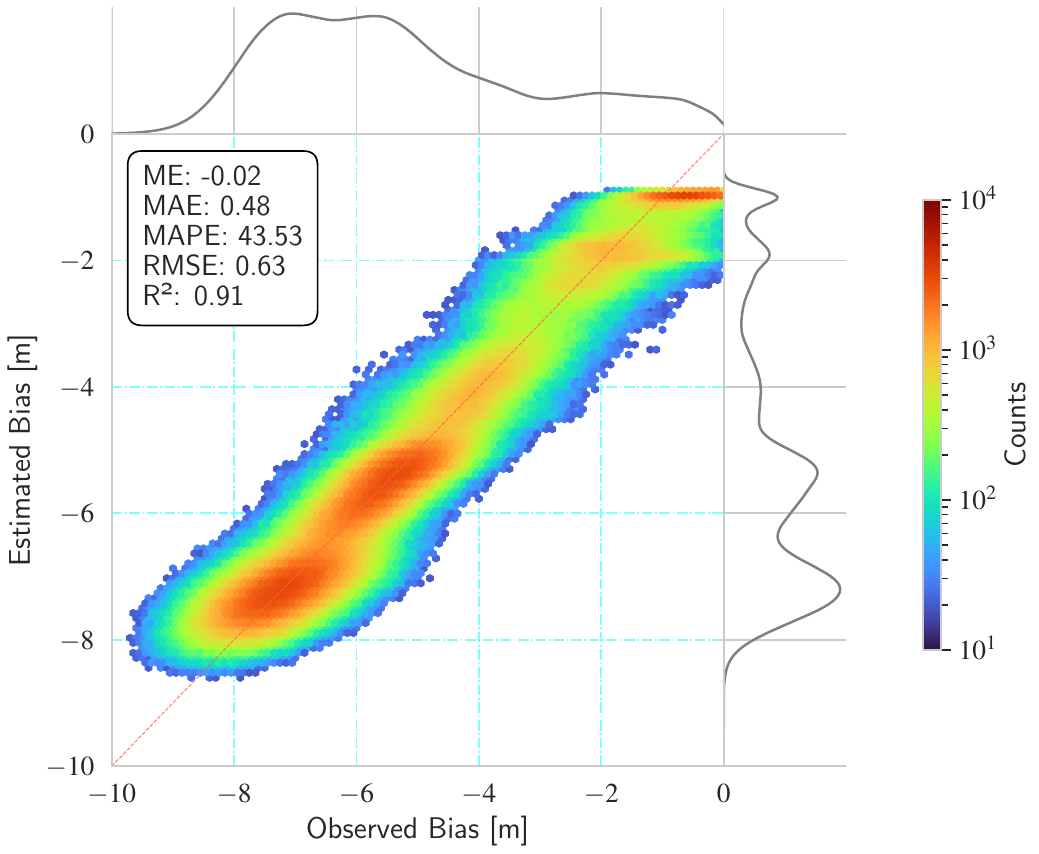}
    \end{minipage}
    \begin{minipage}{0.30\textwidth}
        \centering
        \includegraphics[width=\linewidth]{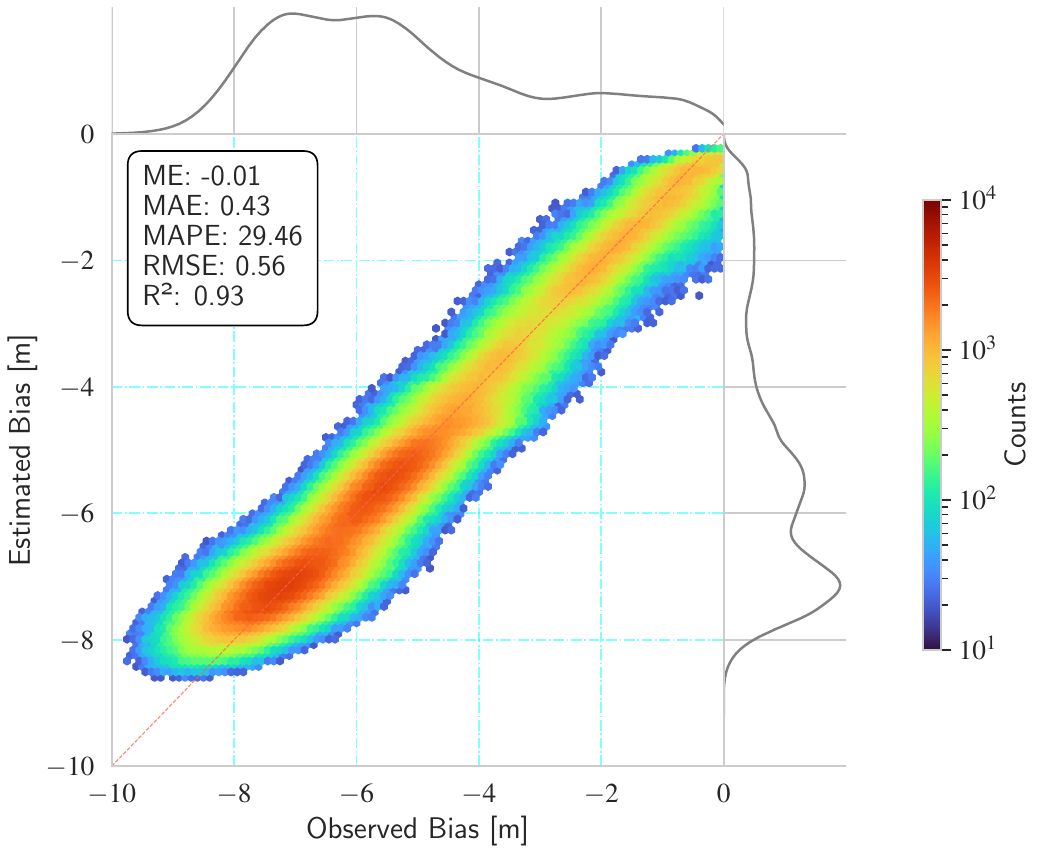}
    \end{minipage}

    \begin{minipage}{0.30\textwidth}
        \centering
        \includegraphics[width=\linewidth]{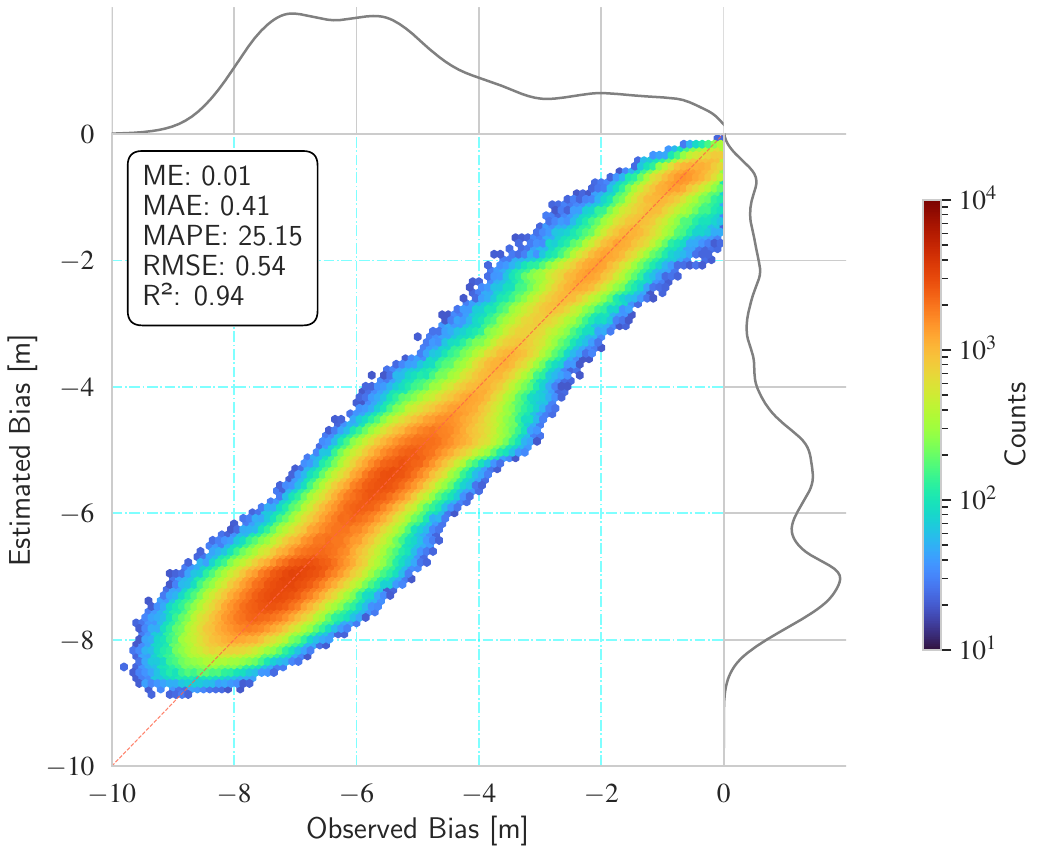}
    \end{minipage}
    \begin{minipage}{0.30\textwidth}
        \centering
        \includegraphics[width=\linewidth]{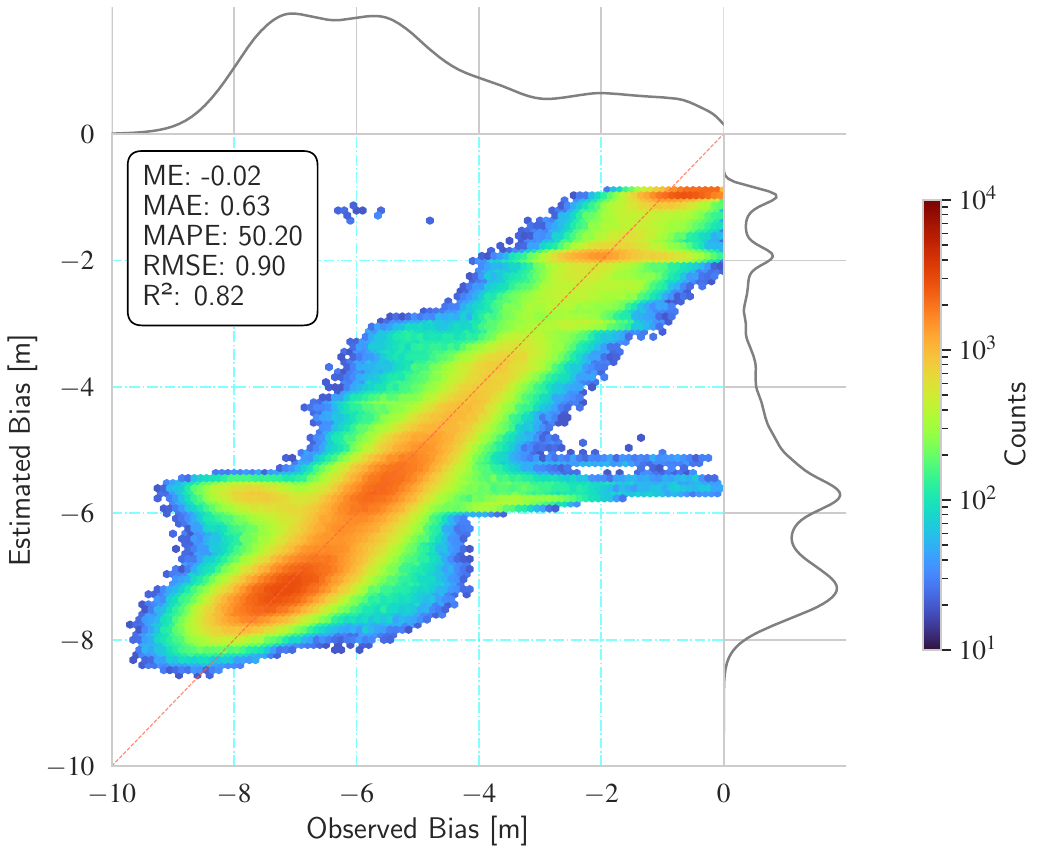}
    \end{minipage}
    \begin{minipage}{0.30\textwidth}
        \centering
        \includegraphics[width=\linewidth]{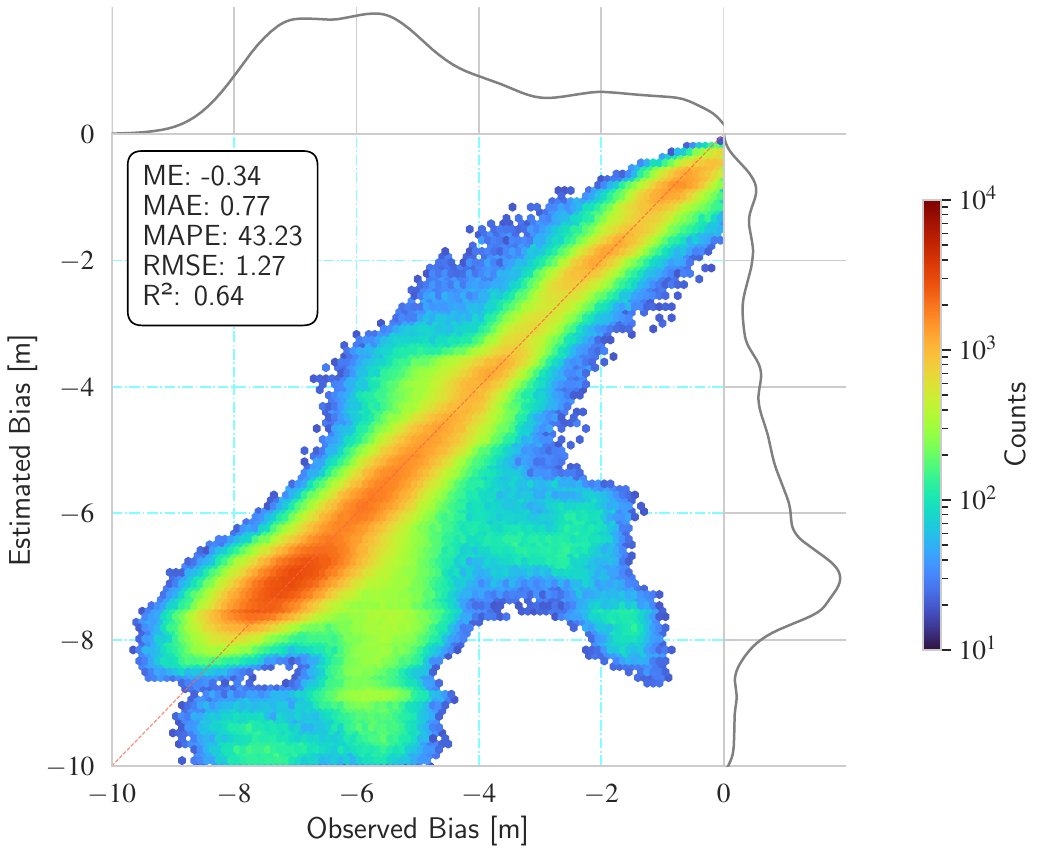}
    \end{minipage}

    \begin{minipage}{0.30\textwidth}
        \centering
        \includegraphics[width=\linewidth]{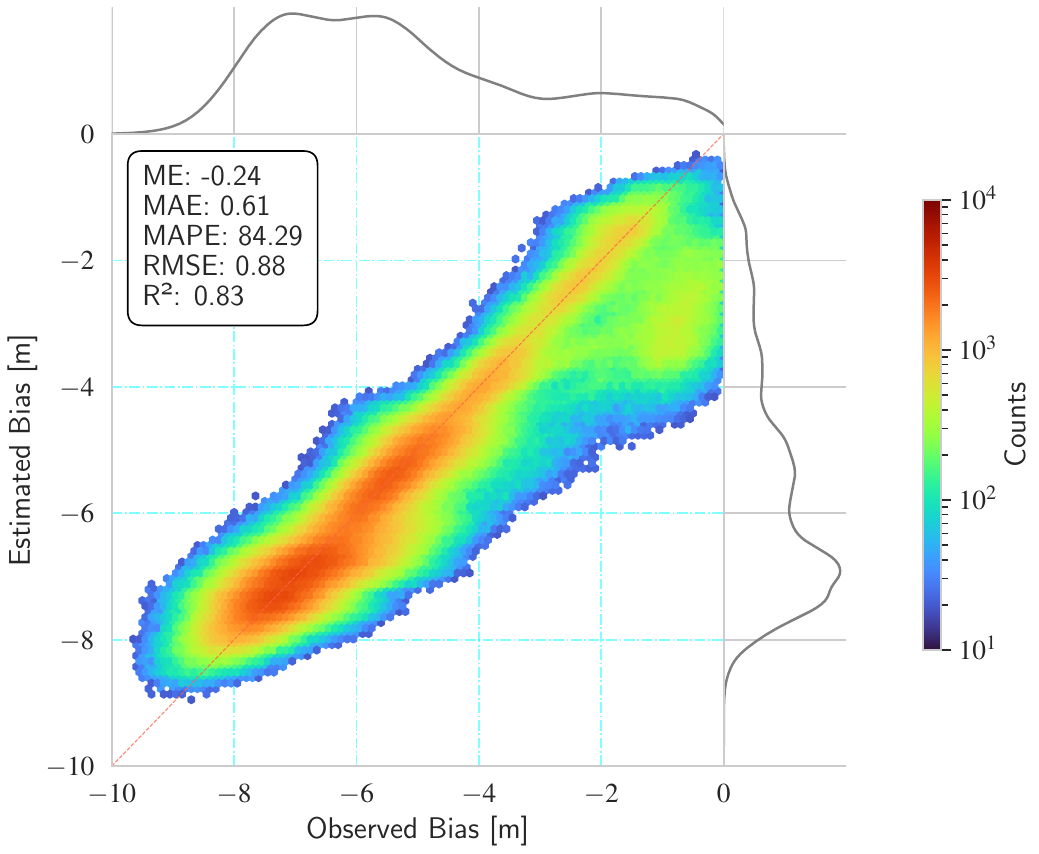}
    \end{minipage}
    \begin{minipage}{0.30\textwidth}
        \centering
        \includegraphics[width=\linewidth]{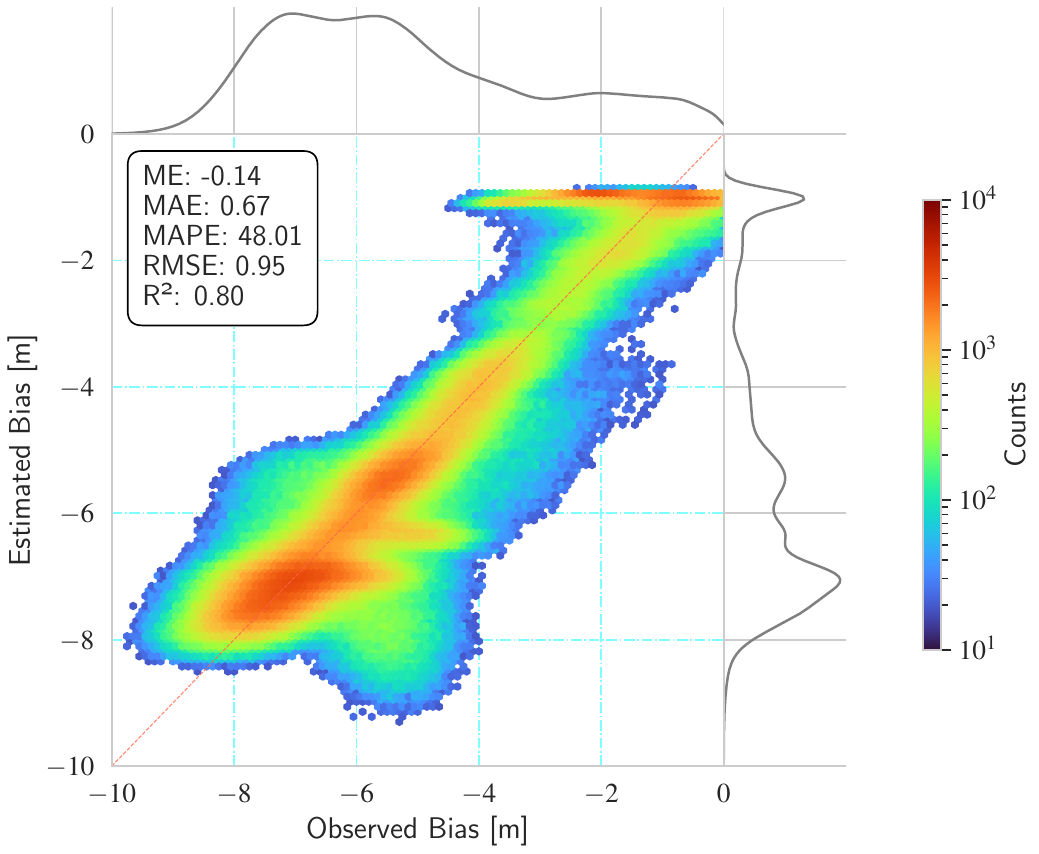}
    \end{minipage}
    \begin{minipage}{0.30\textwidth}
        \centering
        \includegraphics[width=\linewidth]{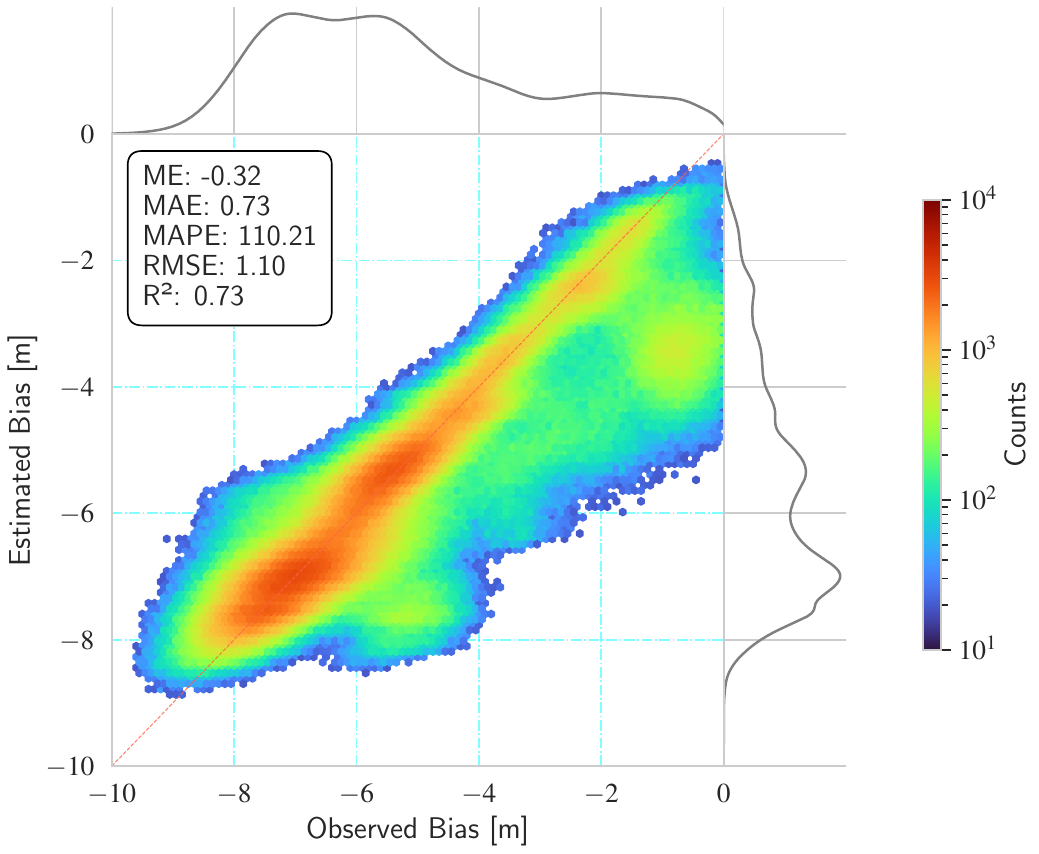}
    \end{minipage}

    \caption{Comparison of model estimations under different training and evaluation conditions. 
    The \textbf{columns} represent different modeling approaches: 
    (Left) Hybrid Model with an Exponential Profile, 
    (Middle) Hybrid Model with a Weibull Profile, 
    (Right) Pure Machine Learning (ML) model using MLP. 
    The \textbf{rows} indicate different training scenarios: 
    (Top) Trained with \emph{All} HoA values, 
    (Middle) \emph{Interpolation} experiment excluding HoA values between 50 and 60 from training, 
    (Bottom) \emph{Extrapolation} experiment excluding HoA values above 70 from training.
    A  comparison with the UV model is provided in Figure~\ref{fig:2d_histogram_pm} of the Supplementary Material .
    }
    
    \label{fig:2D_hist_models}
\end{figure*}

\subsection{Results}
We present the quantitative findings for the different experiments for all the TanDEM-X scenes. The overall performance metrics (ME, MAE, MAPE, RMSE, and $R^2$) are shown in Table~\ref{tab:all_approaches}. It also shows the error distribution statistics (mean \(\mu\) and standard deviation \(\sigma\)) for the uncorrected TanDEM-X DEM, the TanDEM-X DEM corrected by applying the UV model's estimated bias, and the three Hybrid/ML methods, respectively. Results are reported for three different HoA training scenarios: \emph{All}, \emph{Interpolation}, and \emph{Extrapolation}.

The uncorrected DEM exhibits a large negative bias (e.g., \(\mu \approx -5.14\)) with a wide error spread (\(\sigma \approx 2.28\)). 
The UV correction reduces the bias partially (with \(\mu \approx 1.19\)), see Figure~\ref{fig:dem_correction_uv}, but shows high RMSE (2.07 m) and very low \(R^2\) (0.06) as indicated by the 2D histogram in Figure~\ref{fig:2d_histogram_pm} in the Supplementary Material .
This indicates that the penetration bias inverted with a UV model only from the volumetric coherence Eq.~\ref{eq:uv_phase}, is insufficient to capture the actual scattering behavior.

On the other side, the hybrid methods (\emph{Exponential} and \emph{Weibull}) significantly improve the performance. Especially, the Exponential hybrid model achieves the lowest RMSE (as low as 0.52 m) and highest \(R^2\) (0.94) in the \emph{All} scenario. Both the hybrid and the MLP approaches produce comparable results when training on the full HoA range.

\begin{figure}[!htb]
    \centering
    \includegraphics[width=0.95\linewidth]{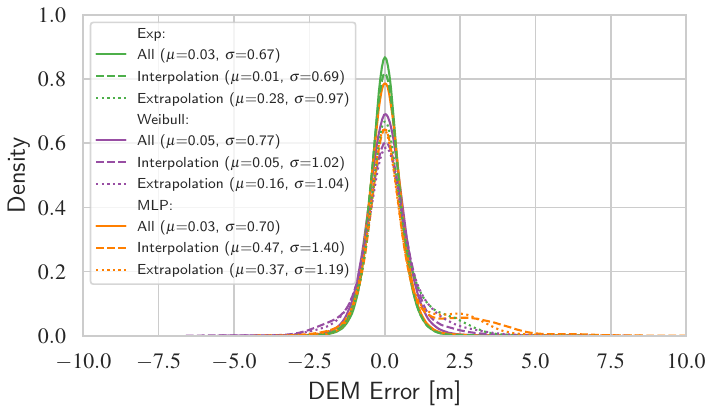}
    \caption{DEM error distributions for the nine modeling approaches 
    (\emph{Exponential}, \emph{Weibull}, and \emph{MLP} under three HoA scenarios) evaluated 
    across all 18 TanDEM-X acquisitions. The mean $\mu$ and standard deviation 
    $\sigma$ are shown in the legend.}
    \label{fig:performance_all}
\end{figure}

Figure~\ref{fig:performance_all} shows the performance of the \emph{Exponential}, \emph{Weibull}, and \emph{MLP} models under three training scenarios. When the entire HoA range is used (\emph{All}), all methods yield tight error distributions centered near zero. Excluding HoA values between 50 and 60 m during training (\emph{Interpolation}) increases errors; here, the Exponential model retains low errors and high $R^2$ compared to the more variable \emph{MLP} and \emph{Weibull} models. Excluding HoA above 70 m during training (\emph{Extrapolation}) leads to a more pronounced performance drop, especially for the \emph{MLP}, while the \emph{Exponential} model remains robust and the \emph{Weibull} model shows moderate degradation.

Figure~\ref{fig:2D_hist_models} illustrates 2D histograms of estimated versus observed penetration bias for the \emph{Exponential}, \emph{Weibull}, and \emph{MLP} models under three HoA training scenarios. In the \emph{All} scenario (top row), all methods cluster tightly around the diagonal line, indicating minimal residual bias. When excluding HoA values between 50 and 60~m during training (\emph{Interpolation}, middle row), the \emph{Exponential} model retains a narrow spread and high correlation with the true bias, whereas the \emph{Weibull} and \emph{MLP} plots exhibit more dispersion. Under the \emph{Extrapolation} scenario (bottom row), which excludes HoA values above 70~m during training, the \emph{MLP} displays a broader scatter, while the \emph{Exponential} model remains comparatively well aligned with the diagonal and the \emph{Weibull} model shows moderate degradation. Overall, these histograms confirm that the \emph{Exponential} model is the most robust to unseen HoA conditions, followed by the \emph{Weibull} and \emph{MLP} approaches. For a spatial visualization of the predicted penetration bias maps, see Figure~\ref{fig:map_penetration_models} in the Supplementary Material.

Figure~\ref{fig:unseen_hoa_distributions} depicts the error distributions for the six models that were trained excluding specific HoA ranges, and then evaluated on unseen HoA scenes. The \emph{Exponential} model consistently exhibits narrower error distributions and lower mean errors, confirming its robust generalization even when key HoA values are missing. The \emph{Weibull} model maintains a near-zero mean under the \emph{Interpolation scenario} but shows a broader spread, and its mean shifts moderately under the \emph{Extrapolation scenario} — though it still avoids large outliers. In contrast, the \emph{MLP} model displays the largest mean shift and widest spread, indicating a higher sensitivity to out-of-distribution HoA conditions. Overall, these results confirm that physically constrained models (particularly the \emph{Exponential} model, followed by the \emph{Weibull} model) are better suited to handling unseen HoA scenarios. See Figure~\ref{fig:2D_hist_models_no_training} in the Supplementary Material  for the corresponding 2D histograms.


\begin{table}[!htb]
    \centering
    \scriptsize  
    \setlength{\tabcolsep}{2.5pt}  
    \caption{Overall performance metrics for the uncorrected DEM and for the DEM corrected by the Physical (Uniform Volume) model and the three Hybrid/ML models (\emph{Exponential}, \emph{Weibull}, \emph{MLP}). Results are shown under the three HoA scenarios for training (\emph{All}, \emph{Interpolation}, \emph{Extrapolation}), using all 18 TanDEM-X scenes. Columns 3--7 show standard error metrics, while the last two columns (\(\mu\) and \(\sigma\)) correspond to the DEM error distribution.}
    \label{tab:all_approaches}
    \label{tab:results_all_with_mu_sigma}
    \begin{tabular}{@{}l l c c c c c c c@{}}
    \toprule
    \textbf{Approach} & \textbf{Scenario} & \textbf{ME} & \textbf{MAE} & \textbf{MAPE} & \textbf{RMSE} & \textbf{R\textsuperscript{2}} & \boldmath{$\mu$} & \boldmath{$\sigma$} \\
    \midrule
    \midrule
    \textbf{Uncorrected} & -- & 
    -- & -- & -- & -- & -- & -5.14 & 2.28 \\
    \midrule
    \textbf{Physical (UV)} & -- & 
    -1.11 & 1.64 & 208.75 & 2.07 & 0.06 & 1.19 & 1.88 \\
    \midrule
    \textbf{\emph{Exponential}} & \textbf{\emph{All}}           
     & -0.01 & 0.40 & 24.24   & 0.52 & 0.94 & 0.03 & 0.67 \\
    \textbf{\emph{Exponential}} & \textbf{\emph{Interpolation}} 
     &  0.01 & 0.41 & 25.15   & 0.54 & 0.94 & 0.01 & 0.69 \\
    \textbf{\emph{Exponential}} & \textbf{\emph{Extrapolation}} 
     & -0.24 & 0.61 & 84.29   & 0.88 & 0.83 & 0.28 & 0.97 \\
    \midrule
    \textbf{\emph{Weibull}} & \textbf{\emph{All}}           
     & -0.02 & 0.48 & 43.53   & 0.63 & 0.91 & 0.05 & 0.77 \\
    \textbf{\emph{Weibull}} & \textbf{\emph{Interpolation}} 
     & -0.02 & 0.63 & 50.20   & 0.90 & 0.82 & 0.05 & 1.02 \\
    \textbf{\emph{Weibull}} & \textbf{\emph{Extrapolation}} 
     & -0.14 & 0.67 & 48.01   & 0.95 & 0.80 & 0.16 & 1.04 \\
    \midrule
    \textbf{\emph{MLP}} & \textbf{\emph{All}}           
     & -0.01 & 0.43 & 29.46   & 0.56 & 0.93 & 0.03 & 0.70 \\
    \textbf{\emph{MLP}} & \textbf{\emph{Interpolation}} 
     & -0.34 & 0.77 & 43.23   & 1.27 & 0.64 & 0.47 & 1.40 \\
    \textbf{\emph{MLP}} & \textbf{\emph{Extrapolation}} 
     & -0.32 & 0.73 & 110.21  & 1.10 & 0.73 & 0.37 & 1.19 \\
    \bottomrule
    \end{tabular}
\end{table}

\begin{figure}[!htb]
    \centering
    \includegraphics[width=0.95\linewidth]{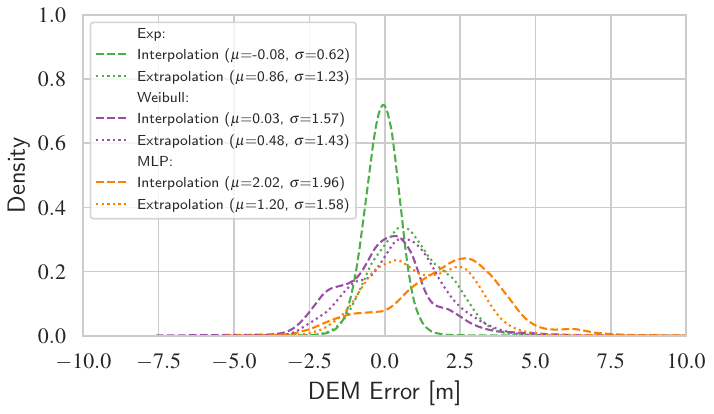}
    \caption{DEM error distributions for the six models that exclude certain 
    HoA ranges during training (\emph{Interpolation} and \emph{Extrapolation}), evaluated 
    on the unseen HoA scenes. The three models trained on all HoA values 
    are omitted here. This isolates how well each method handles 
    out-of-range HoA conditions, with mean $\mu$ and standard deviation 
    $\sigma$ shown in the legend.}
    \label{fig:unseen_hoa_distributions}
\end{figure}
\section{Discussion}
\label{sec:discussion}

\begin{figure}[htb]
    \centering
    \includegraphics[width=0.78\linewidth]{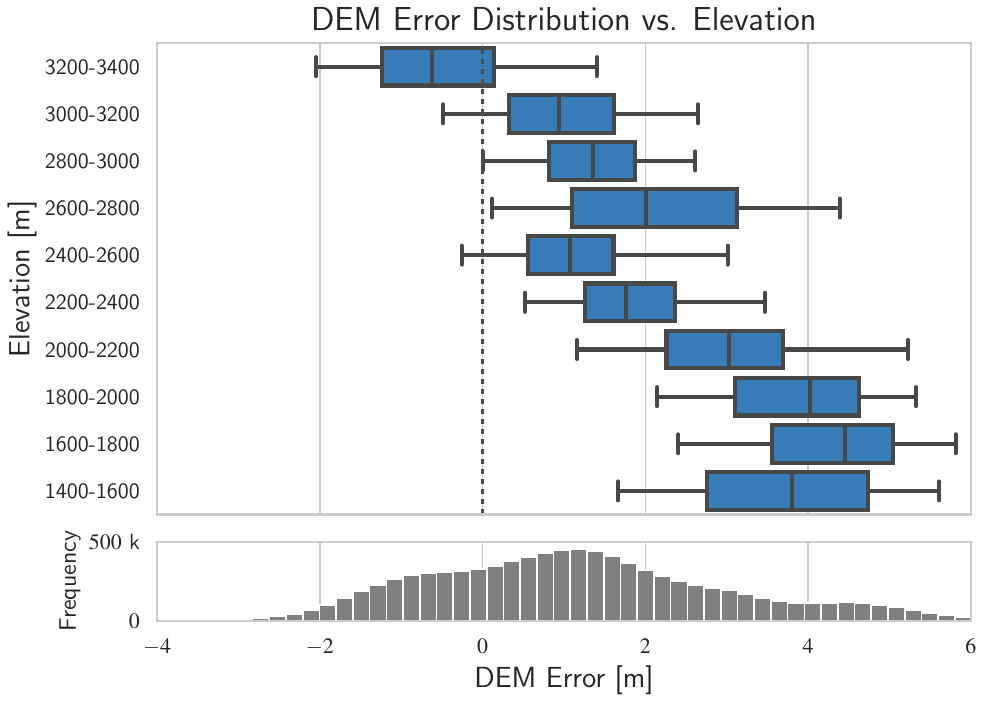}
    \caption{Corrected DEM "using physical (UV) model" error distribution across elevation bins, computed using ATM LiDAR as reference.}
    \label{fig:dem_elevation_error_histogram_pm}
\end{figure}

Our findings indicate that the hybrid approach, particularly the \emph{Exponential model}, stands out as the most effective method for correcting DEM penetration bias. Even though the \emph{Exponential model} is relatively simple — requiring only one parameter — it consistently delivers low RMSE and MAE values along with high \(R^2\) scores across all training scenarios. Our experiments show that the model maintains pronounced robustness under both the \emph{Interpolation scenario} and the \emph{Extrapolation scenario} when evaluated on unseen HoA conditions. This consistent performance is further demonstrated in Figure~\ref{fig:dem_elevation_error_histogram_all} of the supplemental material, where the \emph{Exponential model} (left column) shows stable error distributions across different elevation bins for all scenarios. A small weakness of the \emph{Exponential model} in the \emph{Extrapolation scenario} are unseen HoA at low elevations. The wider error distributions of the \emph{Weibull model} are evident across all elevations, whereas the errors of the \emph{MLP model} are at elevation bins dominated by unseen HoAs. 
In contrast, the elevation-dependent error of the pure physical UV model in Figure~\ref{fig:dem_elevation_error_histogram_pm} matches the theoretical expectation that the underlying uniform volume assumption fits best at the highest elevations, whereas the subsurface structure becomes increasingly more heterogeneous towards lower elevations, which results in larger model errors.

When the full range of HoA values is available during training, both hybrid and pure ML methods perform comparably. This suggests that penetration bias estimation with ML achieves high accuracies when using balanced training data, independent of the actual ML method, which is in line with literature \cite{abdullahi_estimating_2019,campos_potential_2024}. However, excluding key HoA ranges reveals the advantage of incorporating physical constraints. The \emph{Exponential model} leverages its physically informed vertical scattering profile to regularize parameter estimation, which helps prevent overfitting and ensures stable performance even with limited training data. In contrast, the \emph{Weibull model}, which requires estimating two parameters, exhibits more significant variability due to its inherent ill-posedness; its additional parameter (\(k_w\)) appears to correlate with HoA, which it theoretically should not do, leading to moderate performance degradation under the \emph{Extrapolation scenario}. The \emph{MLP model} shows the largest increase in error variance when tested on unseen HoA conditions, indicating a higher sensitivity to out-of-distribution data.
\section{Conclusion}
\label{sec:conclusion}
This study introduces a hybrid framework that combines physical modeling with data-driven techniques to correct the DEM penetration bias in TanDEM-X InSAR data over Greenland. Our experiments show that hybrid and pure ML methods perform adequately when the full range of HoA values is available. However, the hybrid approach outperforms pure ML when key HoA ranges are missing during training. In particular, the \emph{Exponential model} — with its simple one-parameter design — consistently achieves the lowest error metrics and highest stability, even under challenging \emph{Interpolation} and \emph{Extrapolation} scenarios.

The advantages of our hybrid approach are threefold: (1) the \emph{Exponential model} consistently produces the lowest errors and highest stability across all HoA training scenarios, (2) incorporating physical constraints reduces the need for a highly diverse training dataset while enhancing generalization to unseen acquisition geometries, and (3) embedding the physically informed vertical scattering profile improves accuracy and interpretability, making it a robust solution for DEM bias correction.

Our findings demonstrate that embedding physical constraints into the learning process enhances performance and helps overcome challenges such as incomplete HoA coverage (i.e., limited diversity in acquisition geometry), a common issue in operational scenarios. Moreover, our hybrid framework is uniquely suited for integrating multi-modal and multi-sensor data into the physical model — a task that is typically challenging with traditional methods. This flexibility facilitates adaptation to other satellite missions (e.g., ESA's Sentinels and Harmony) operating at different wavelengths and imaging conditions, improving performance even when high-quality reference data are scarce.

\clearpage
{
    \small
    \bibliographystyle{ieeenat_fullname}
    \bibliography{references}

\begin{thebibliography}{32}
\providecommand{\natexlab}[1]{#1}
\providecommand{\url}[1]{\texttt{#1}}
\expandafter\ifx\csname urlstyle\endcsname\relax
  \providecommand{\doi}[1]{doi: #1}\else
  \providecommand{\doi}{doi: \begingroup \urlstyle{rm}\Url}\fi

\bibitem[Abdullahi et~al.(2019)Abdullahi, Wessel, Huber, Wendleder, Roth, and
  Kuenzer]{abdullahi_estimating_2019}
Sahra Abdullahi, Birgit Wessel, Martin Huber, Anna Wendleder, Achim Roth, and
  Claudia Kuenzer.
\newblock Estimating {Penetration}-{Related} {X}-{Band} {InSAR} {Elevation}
  {Bias}: {A} {Study} over the {Greenland} {Ice} {Sheet}.
\newblock \emph{Remote Sensing}, 11\penalty0 (24):\penalty0 2903, 2019.

\bibitem[Alexandrov et~al.(2018)Alexandrov, McMichael, and
  Beyer]{alexandrov_icebridge_2018}
Oleg Alexandrov, Scott McMichael, and Ross Beyer.
\newblock Icebridge dms l3 ames stereo pipeline photogrammetric dem, version 1,
  2018.

\bibitem[Askne et~al.(1997)Askne, Dammert, Ulander, and
  Smith]{askne_c-band_1997}
J.I.H. Askne, P.B.G. Dammert, L.M.H. Ulander, and G. Smith.
\newblock C-band repeat-pass interferometric {SAR} observations of the forest.
\newblock \emph{IEEE Transactions on Geoscience and Remote Sensing},
  35\penalty0 (1):\penalty0 25--35, 1997.

\bibitem[Bamler and Hartl(1998)]{bamler_synthetic_1998}
Richard Bamler and Philipp Hartl.
\newblock Synthetic aperture radar interferometry.
\newblock \emph{Inverse Problems}, 14\penalty0 (4):\penalty0 R1, 1998.

\bibitem[Becker~Campos et~al.(2024)Becker~Campos, Rizzoli, Bueso-Bello, and
  Braun]{becker_campos_unsupervised_2024}
Alexandre Becker~Campos, Paola Rizzoli, Jose~Luis Bueso-Bello, and Matthias
  Braun.
\newblock An {Unsupervised} {Deep} {Learning} {Approach} for {Monitoring} the
  {Snow} {Facies} of the {Greenland} {Ice} {Sheet} with {InSAR} {TanDEM}-{X}
  {Data}.
\newblock In \emph{{EUSAR} 2024; 15th {European} {Conference} on {Synthetic}
  {Aperture} {Radar}}, pages 175--180, 2024.

\bibitem[Campos et~al.(2024)Campos, Braun, and Rizzoli]{campos_potential_2024}
Alexandre~Becker Campos, Matthias Braun, and Paola Rizzoli.
\newblock On the {Potential} of {Bistatic} {Insar} {Features} for {Monitoring}
  {Ice} {Sheets} {Properties} and {Estimating} {Surface} {Elevation} {Bias}.
\newblock In \emph{{IGARSS} 2024 - 2024 {IEEE} {International} {Geoscience} and
  {Remote} {Sensing} {Symposium}}, pages 143--146, Athens, Greece, 2024. IEEE.

\bibitem[Cloude(2009)]{cloude_polarisation_2009}
Shane Cloude.
\newblock \emph{Polarisation: {Applications} in {Remote} {Sensing}}.
\newblock Oxford University Press, 2009.

\bibitem[Cloude and Papathanassiou(1998)]{cloude_polarimetric_1998}
S.R. Cloude and K.P. Papathanassiou.
\newblock Polarimetric {SAR} interferometry.
\newblock \emph{IEEE Transactions on Geoscience and Remote Sensing},
  36\penalty0 (5):\penalty0 1551--1565, 1998.

\bibitem[Dall(2007)]{dall_insar_2007}
Jrgen Dall.
\newblock {InSAR} {Elevation} {Bias} {Caused} by {Penetration} {Into} {Uniform}
  {Volumes}.
\newblock \emph{IEEE Transactions on Geoscience and Remote Sensing},
  45\penalty0 (7):\penalty0 2319--2324, 2007.

\bibitem[Dall et~al.(2001)Dall, Madsen, Keller, and
  Forsberg]{dall_topography_2001}
Jørgen Dall, Søren~Nørvang Madsen, Kristian Keller, and Rene Forsberg.
\newblock Topography and penetration of the {Greenland} {Ice} {Sheet} measured
  with {Airborne} {SAR} {Interferometry}.
\newblock \emph{Geophysical Research Letters}, 28\penalty0 (9):\penalty0
  1703--1706, 2001.

\bibitem[{European Space Agency} and {Airbus}(2022)]{noauthor_copernicus_2022}
{European Space Agency} and {Airbus}.
\newblock Copernicus {DEM}, 2022.

\bibitem[Fan et~al.(2022)Fan, Ke, and Shen]{fan_new_2022}
Yubin Fan, Chang-Qing Ke, and Xiaoyi Shen.
\newblock A new {Greenland} digital elevation model derived from {ICESat}-2
  during 2018–2019.
\newblock \emph{Earth System Science Data}, 14\penalty0 (2):\penalty0 781--794,
  2022.

\bibitem[Fischer et~al.(2019)Fischer, Jäger, Papathanassiou, and
  Hajnsek]{fischer_modeling_2019}
Georg Fischer, Marc Jäger, Konstantinos~P. Papathanassiou, and Irena Hajnsek.
\newblock Modeling the {Vertical} {Backscattering} {Distribution} in the
  {Percolation} {Zone} of the {Greenland} {Ice} {Sheet} {With} {SAR}
  {Tomography}.
\newblock \emph{IEEE Journal of Selected Topics in Applied Earth Observations
  and Remote Sensing}, 12\penalty0 (11):\penalty0 4389--4405, 2019.

\bibitem[Fischer et~al.(2020)Fischer, Papathanassiou, and
  Hajnsek]{fischer_modeling_2020}
Georg Fischer, Konstantinos~P. Papathanassiou, and Irena Hajnsek.
\newblock Modeling and {Compensation} of the {Penetration} {Bias} in {InSAR}
  {DEMs} of {Ice} {Sheets} at {Different} {Frequencies}.
\newblock \emph{IEEE Journal of Selected Topics in Applied Earth Observations
  and Remote Sensing}, 13:\penalty0 2698--2707, 2020.

\bibitem[Gatelli et~al.(1994)Gatelli, Monti~Guamieri, Parizzi, Pasquali, Prati,
  and Rocca]{gatelli_wavenumber_1994}
F. Gatelli, A. Monti~Guamieri, F. Parizzi, P. Pasquali, C. Prati, and F. Rocca.
\newblock The wavenumber shift in {SAR} interferometry.
\newblock \emph{IEEE Transactions on Geoscience and Remote Sensing},
  32\penalty0 (4):\penalty0 855--865, 1994.

\bibitem[Gibson et~al.(2021)Gibson, Chapman, Altinok, Delle~Monache, DeFlorio,
  and Waliser]{gibson_training_2021}
Peter~B. Gibson, William~E. Chapman, Alphan Altinok, Luca Delle~Monache,
  Michael~J. DeFlorio, and Duane~E. Waliser.
\newblock Training machine learning models on climate model output yields
  skillful interpretable seasonal precipitation forecasts.
\newblock \emph{Communications Earth \& Environment}, 2\penalty0 (1):\penalty0
  1--13, 2021.

\bibitem[Hagberg et~al.(1995)Hagberg, Ulander, and
  Askne]{hagberg_repeat-pass_1995}
Jan~O. Hagberg, Lars~M.H. Ulander, and Jan Askne.
\newblock Repeat-pass {SAR} interferometry over forested terrain.
\newblock \emph{IEEE Transactions on Geoscience and Remote Sensing},
  33\penalty0 (2):\penalty0 331--340, 1995.

\bibitem[Karniadakis et~al.(2021)Karniadakis, Kevrekidis, Lu, Perdikaris, Wang,
  and Yang]{karniadakis_physics-informed_2021}
George~Em Karniadakis, Ioannis~G. Kevrekidis, Lu Lu, Paris Perdikaris, Sifan
  Wang, and Liu Yang.
\newblock Physics-informed machine learning.
\newblock \emph{Nature Reviews Physics}, 3\penalty0 (6):\penalty0 422--440,
  2021.

\bibitem[Krasnopolsky and Fox-Rabinovitz(2006)]{krasnopolsky_complex_2006}
Vladimir~M. Krasnopolsky and Michael~S. Fox-Rabinovitz.
\newblock Complex hybrid models combining deterministic and machine learning
  components for numerical climate modeling and weather prediction.
\newblock \emph{Neural Networks}, 19\penalty0 (2):\penalty0 122--134, 2006.

\bibitem[Krieger et~al.(2007)Krieger, Moreira, Fiedler, Hajnsek, Werner,
  Younis, and Zink]{krieger_tandem-x_2007}
Gerhard Krieger, Alberto Moreira, Hauke Fiedler, Irena Hajnsek, Marian Werner,
  Marwan Younis, and Manfred Zink.
\newblock {TanDEM}-{X}: {A} {Satellite} {Formation} for {High}-{Resolution}
  {SAR} {Interferometry}.
\newblock \emph{IEEE Transactions on Geoscience and Remote Sensing},
  45\penalty0 (11):\penalty0 3317--3341, 2007.

\bibitem[Kurz et~al.(2022)Kurz, De~Gersem, Galetzka, Klaedtke, Liebsch,
  Loukrezis, Russenschuck, and Schmidt]{Kurz:2022gtq}
Stefan Kurz, Herbert De~Gersem, Armin Galetzka, Andreas Klaedtke, Melvin
  Liebsch, Dimitrios Loukrezis, Stephan Russenschuck, and Manuel Schmidt.
\newblock Hybrid modeling: towards the next level of scientific computing in
  engineering.
\newblock \emph{Journal of Mathematics in Industry}, 12\penalty0 (1):\penalty0
  8, 2022.

\bibitem[Mansour et~al.(2024)Mansour, Fischer, Hänsch, Hajnsek, and
  Papathanassiou]{mansour_correction_2024}
Islam Mansour, Georg Fischer, Ronny Hänsch, Irena Hajnsek, and Konstantinos
  Papathanassiou.
\newblock Correction of {The} {Penetration} {Bias} for {InSAR} {DEM} {Via}
  {Synergetic} {AI}-{Physical} {Modeling}: {A} {Greenland} {Case} {Study}.
\newblock In \emph{{IGARSS} 2024 - 2024 {IEEE} {International} {Geoscience} and
  {Remote} {Sensing} {Symposium}}, pages 138--142, 2024.

\bibitem[Mansour et~al.(2025)Mansour, Papathanassiou, Hänsch, and
  Hajnsek]{mansour_hybrid_2025}
Islam Mansour, Konstantinos Papathanassiou, Ronny Hänsch, and Irena Hajnsek.
\newblock Hybrid {Machine} {Learning} {Forest} {Height} {Estimation} {From}
  {TanDEM}-{X} {InSAR}.
\newblock \emph{IEEE Transactions on Geoscience and Remote Sensing},
  63:\penalty0 1--11, 2025.

\bibitem[Martone et~al.(2015)Martone, Bräutigam, and
  Krieger]{martone_quantization_2015}
Michele Martone, Benjamin Bräutigam, and Gerhard Krieger.
\newblock Quantization {Effects} in {TanDEM}-{X} {Data}.
\newblock \emph{IEEE Transactions on Geoscience and Remote Sensing},
  53\penalty0 (2):\penalty0 583--597, 2015.

\bibitem[Mätzler(1987)]{matzler_applications_1987}
Christian Mätzler.
\newblock Applications of the interaction of microwaves with the natural snow
  cover.
\newblock \emph{Remote Sensing Reviews}, 2\penalty0 (2):\penalty0 259--387,
  1987.

\bibitem[Rignot et~al.(2001)Rignot, Echelmeyer, and
  Krabill]{rignot_penetration_2001}
Eric Rignot, Keith Echelmeyer, and William Krabill.
\newblock Penetration depth of interferometric synthetic-aperture radar signals
  in snow and ice.
\newblock \emph{Geophysical Research Letters}, 28\penalty0 (18):\penalty0
  3501--3504, 2001.

\bibitem[Rosen et~al.(2000)Rosen, Hensley, Joughin, Li, Madsen, Rodriguez, and
  Goldstein]{rosen_synthetic_2000}
P.A. Rosen, S. Hensley, I.R. Joughin, F.K. Li, S.N. Madsen, E. Rodriguez, and
  R.M. Goldstein.
\newblock Synthetic aperture radar interferometry.
\newblock \emph{Proceedings of the IEEE}, 88\penalty0 (3):\penalty0 333--382,
  2000.

\bibitem[Sharma et~al.(2013)Sharma, Hajnsek, Papathanassiou, and
  Moreira]{sharma_estimation_2013}
Jayanti~J. Sharma, Irena Hajnsek, Konstantinos~P. Papathanassiou, and Alberto
  Moreira.
\newblock Estimation of {Glacier} {Ice} {Extinction} {Using}
  {Long}-{Wavelength} {Airborne} {Pol}-{InSAR}.
\newblock \emph{IEEE Transactions on Geoscience and Remote Sensing},
  51\penalty0 (6):\penalty0 3715--3732, 2013.

\bibitem[Weber~Hoen and Zebker(2000)]{weber_hoen_penetration_2000}
E. Weber~Hoen and H.A. Zebker.
\newblock Penetration depths inferred from interferometric volume decorrelation
  observed over the {Greenland} {Ice} {Sheet}.
\newblock \emph{IEEE Transactions on Geoscience and Remote Sensing},
  38\penalty0 (6):\penalty0 2571--2583, 2000.

\bibitem[Weisz(2020)]{weisz_hybrid_2020}
Gjermund Weisz.
\newblock Hybrid {Modeling} - {Combining} {Physics} and {Machine} {Learning} to
  {Understand} {Multiphase} {Transient} {Flow}, 2020.

\bibitem[Wessel et~al.(2016)Wessel, Bertram, Gruber, Bemm, and
  Dech]{wessel_new_2016}
B. Wessel, A. Bertram, A. Gruber, S. Bemm, and S. Dech.
\newblock A {NEW} {HIGH}-{RESOLUTION} {ELEVATION} {MODEL} {OF} {GREENLAND}
  {DERIVED} {FROM} {TANDEM}-{X}.
\newblock \emph{ISPRS Annals of the Photogrammetry, Remote Sensing and Spatial
  Information Sciences}, III-7:\penalty0 9--16, 2016.

\bibitem[Zebker and Villasenor(1992)]{zebker_decorrelation_1992}
H.A. Zebker and J. Villasenor.
\newblock Decorrelation in interferometric radar echoes.
\newblock \emph{IEEE Transactions on Geoscience and Remote Sensing},
  30\penalty0 (5):\penalty0 950--959, 1992.

\end{thebibliography}
}

\clearpage
\setcounter{page}{1}
\maketitlesupplementary

\section{Definitions of Vertical Wavenumber and HoA}
\label{sec:hoa_supplement}
In our analysis of InSAR data, two key parameters describe how heights (i.e. depths) are calculated from the observed or modelled interferometric phase: the vertical wavenumber \(\kappa_z\) and the Height of Ambiguity (HoA). 
Physically, \(\kappa_z\) depends on the dielectric properties of snow and ice and has to account for the refraction and change in wave propagation speed in the medium, which leads to a vertical wavenumber in the (snow and ice) volume $\kappa_{z_{\mathrm{Vol}}}$. However, TanDEM-X DEMs are produced under the \emph{free-space} assumption, so the penetration bias correction must also adopt the \emph{free-space} vertical wavenumber $\kappa_z$ as~\cite{sharma_estimation_2013}
\begin{equation}
    \tag{S1}
    \label{eq:kappa_z_def}
    \kappa_z = \frac{4\pi}{\lambda}\frac{\Delta \theta_i}{\sin \theta_i},
\end{equation}
where \(\lambda\) denotes the radar wavelength, \(\theta_i\) is the incidence angle, and \(\Delta \theta_i\) is the baseline-induced angular shift. 

The Height of Ambiguity (HoA), which quantifies the phase-to-height sensitivity by representing the elevation difference corresponding to a full \(2\pi\) interferometric phase cycle, is defined as
\begin{equation}
    \tag{S2}
    \label{eq:HoA_def}
    \mathrm{HoA} = \frac{2\pi}{\kappa_z}.
\end{equation}

\section{Metric Definitions}
\label{sec:metrics_supplement}
We evaluate model performance using standard metrics that assess bias, error magnitude, and goodness-of-fit. The following equations define these metrics:

\begin{equation}
    \tag{S3}
    \label{eq:S3}
    \text{ME} = \frac{1}{n} \sum_{i=1}^{n} (\hat{y}_i - y_i),
\end{equation}

\begin{equation}
    \tag{S4}
    \label{eq:S4}
    \text{MAE} = \frac{1}{n} \sum_{i=1}^{n} \left| \hat{y}_i - y_i \right|,
\end{equation}

\begin{equation}
    \tag{S5}
    \label{eq:S5}
    \text{MAPE} = \frac{100}{n} \sum_{i=1}^{n} \left| \frac{\hat{y}_i - y_i}{y_i} \right|,
\end{equation}

\begin{equation}
    \tag{S6}
    \label{eq:S6}
    \text{RMSE} = \sqrt{\frac{1}{n} \sum_{i=1}^{n} (\hat{y}_i - y_i)^2},
\end{equation}

\begin{equation}
    \tag{S7}
    \label{eq:S7}
    R^2 = 1 - \frac{\sum_{i=1}^{n} (\hat{y}_i - y_i)^2}{\sum_{i=1}^{n} (y_i - \bar{y})^2},
\end{equation}

\begin{equation}
    \tag{S8}
    \label{eq:S8}
    \mu = \frac{1}{n} \sum_{i=1}^{n} e_i,
\end{equation}

\begin{equation}
    \tag{S9}
    \label{eq:S9}
    \sigma = \sqrt{\frac{1}{n} \sum_{i=1}^{n} \left( e_i - \mu \right)^2},
\end{equation}

where \(\hat{y}_i\) represents the predicted values, \(y_i\) denotes the corresponding reference values, and \(\bar{y}\) is the mean of the reference values. Additionally, \(e_i = \hat{h}_{\text{InSAR}} - h_{\text{ref}}\) denotes the DEM error before or after applying bias correction. The mean error \(\mu\) provides insight into systematic bias, while the standard deviation \(\sigma\) captures the variability of the DEM error.

\begin{figure}[htb]
    \centering
    \includegraphics[width=0.78\linewidth]{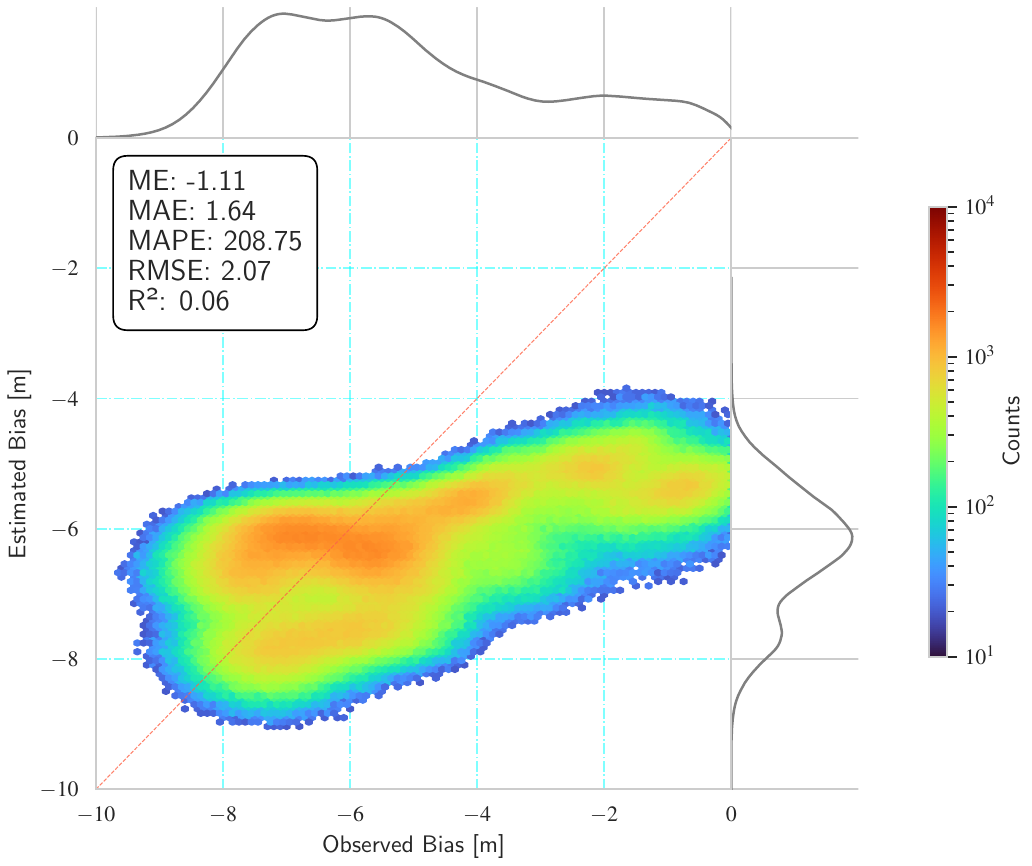}
    \caption{2D histograms of estimated versus observed penetration bias for the Physical (UV) model}
    \label{fig:2d_histogram_pm}
\end{figure}

\begin{figure*}[htb]
    \centering
    \begin{subfigure}[b]{0.95\textwidth}
        \includegraphics[width=\linewidth]{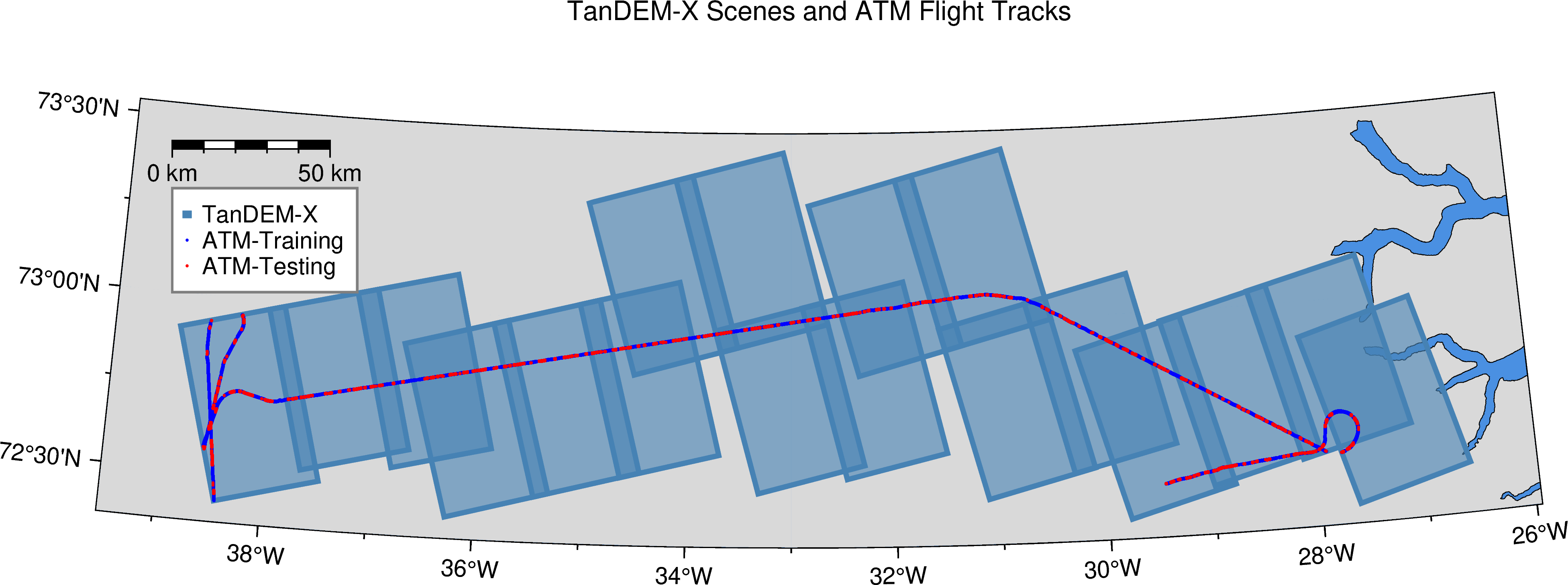}
        \caption{All scenario}
    \end{subfigure}
    \vfill
    \begin{subfigure}[b]{0.95\textwidth}
        \includegraphics[width=\linewidth]{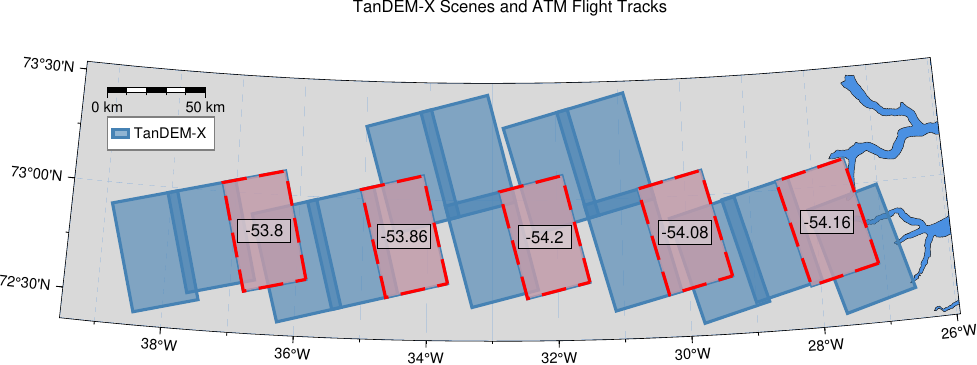}
        \caption{Interpolation scenario}
    \end{subfigure}
    \vfill
    \begin{subfigure}[b]{0.95\textwidth}
        \includegraphics[width=\linewidth]{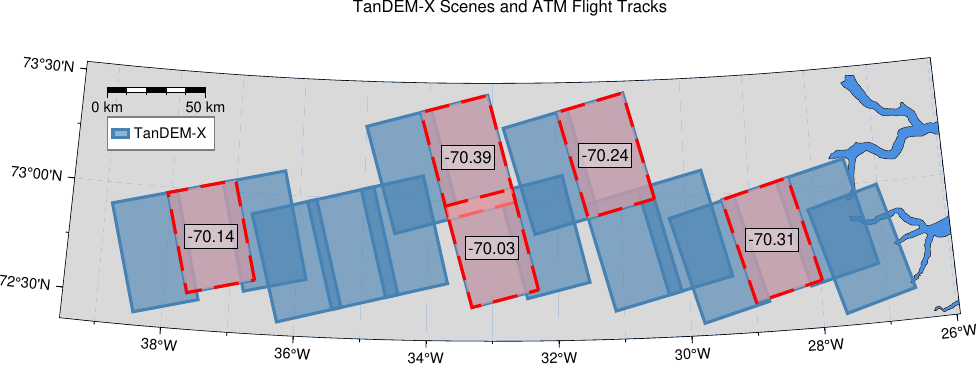}
        \caption{Extrapolation scenario}
    \end{subfigure}
    \caption{Overview of the TanDEM-X scenes (blue) and ATM flight tracks (red/blue) used for each training scenario. The ATM flight lines are split into training (blue) and testing (red) segments, ensuring 
    coverage of different surface conditions for model evaluation.
    Scenes outlined in red are excluded from training under the specified HoA range. 
    (a) \emph{All}: uses every scene, 
    (b) \emph{Interpolation}: excludes HoA in [50,\,60]\,m, 
    (c) \emph{Extrapolation}: excludes HoA above 70\,m.
    }
    \label{fig:hoa_scenarios_supplement}
\end{figure*}

\begin{figure*}[!htb]
        \centering
        \begin{minipage}{0.32\textwidth}
            \centering 
            \includegraphics[width=\linewidth]{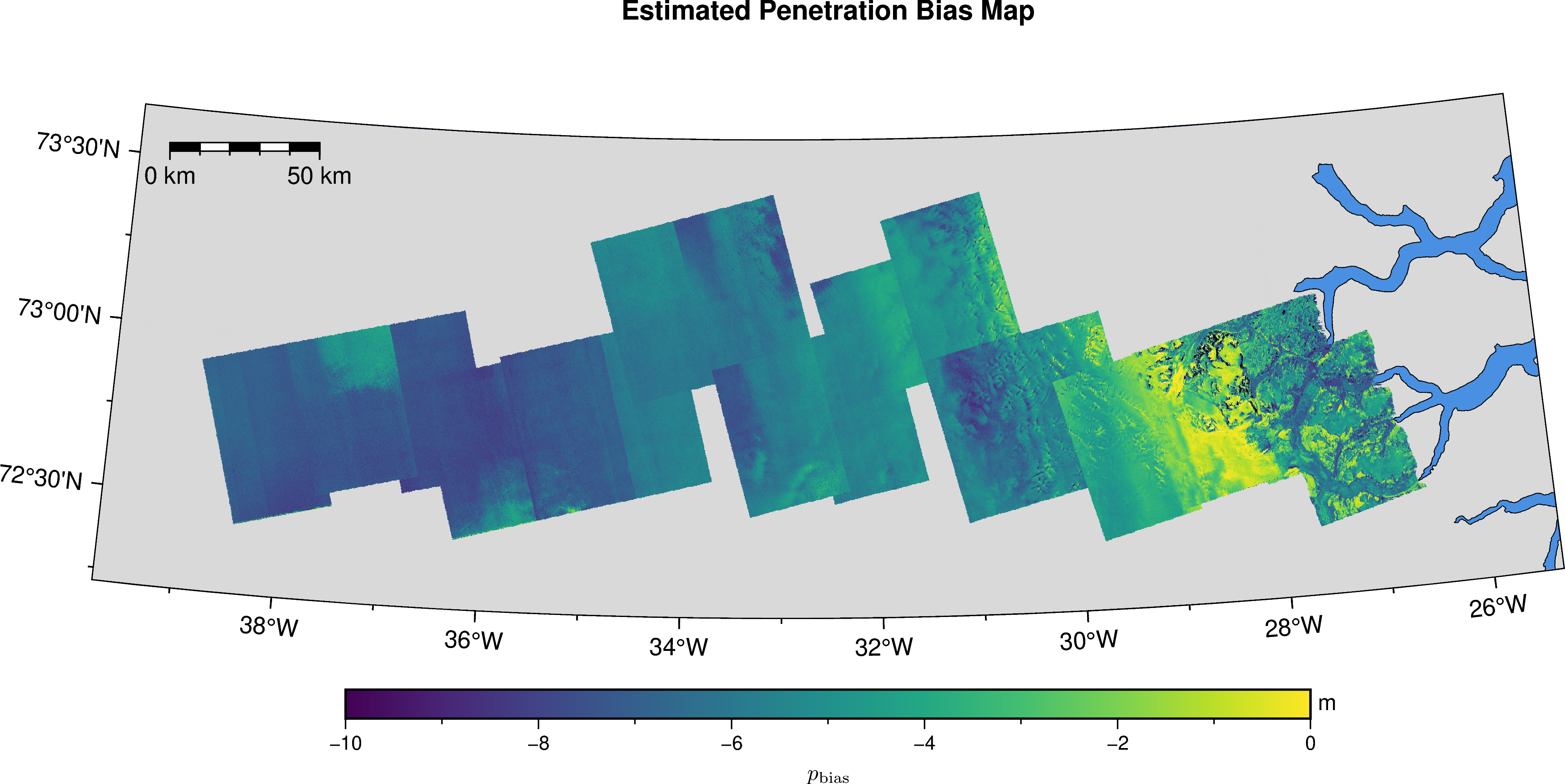}
        \end{minipage}
        \begin{minipage}{0.32\textwidth}
            \centering
            \includegraphics[width=\linewidth]{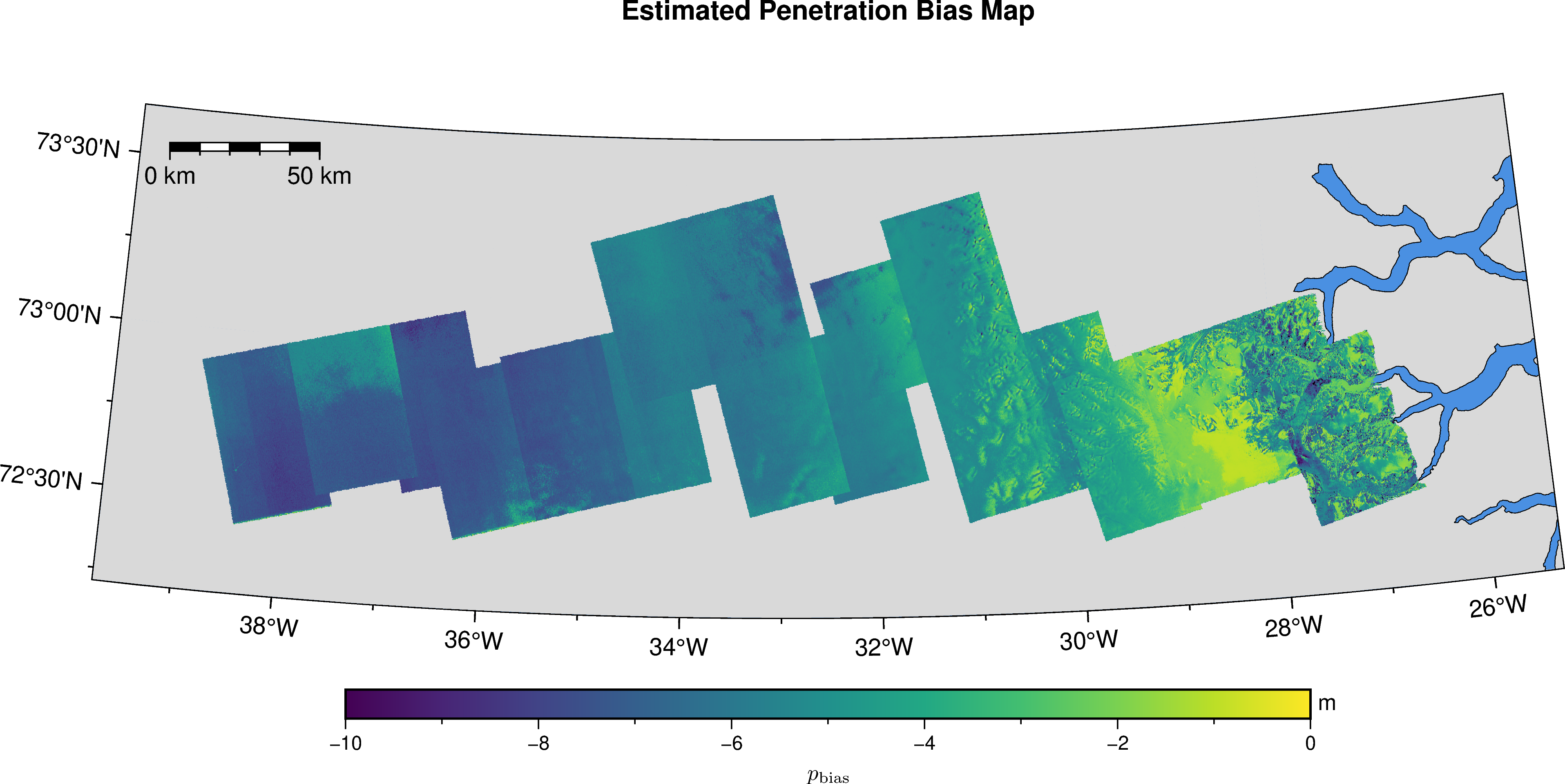}
        \end{minipage}
        \begin{minipage}{0.32\textwidth}
            \centering
            \includegraphics[width=\linewidth]{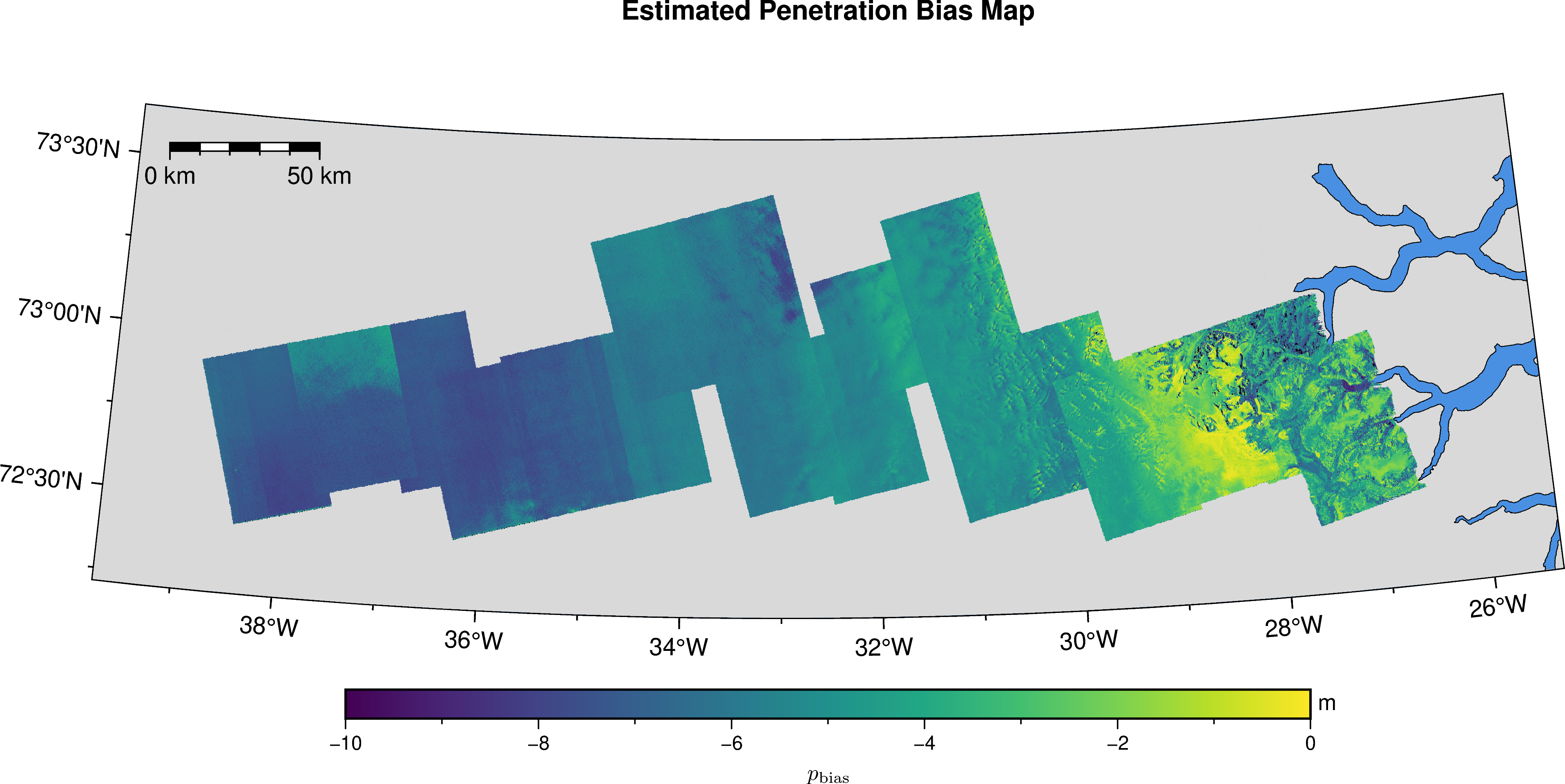}
        \end{minipage}
    
        \begin{minipage}{0.32\textwidth}
            \centering
            \includegraphics[width=\linewidth]{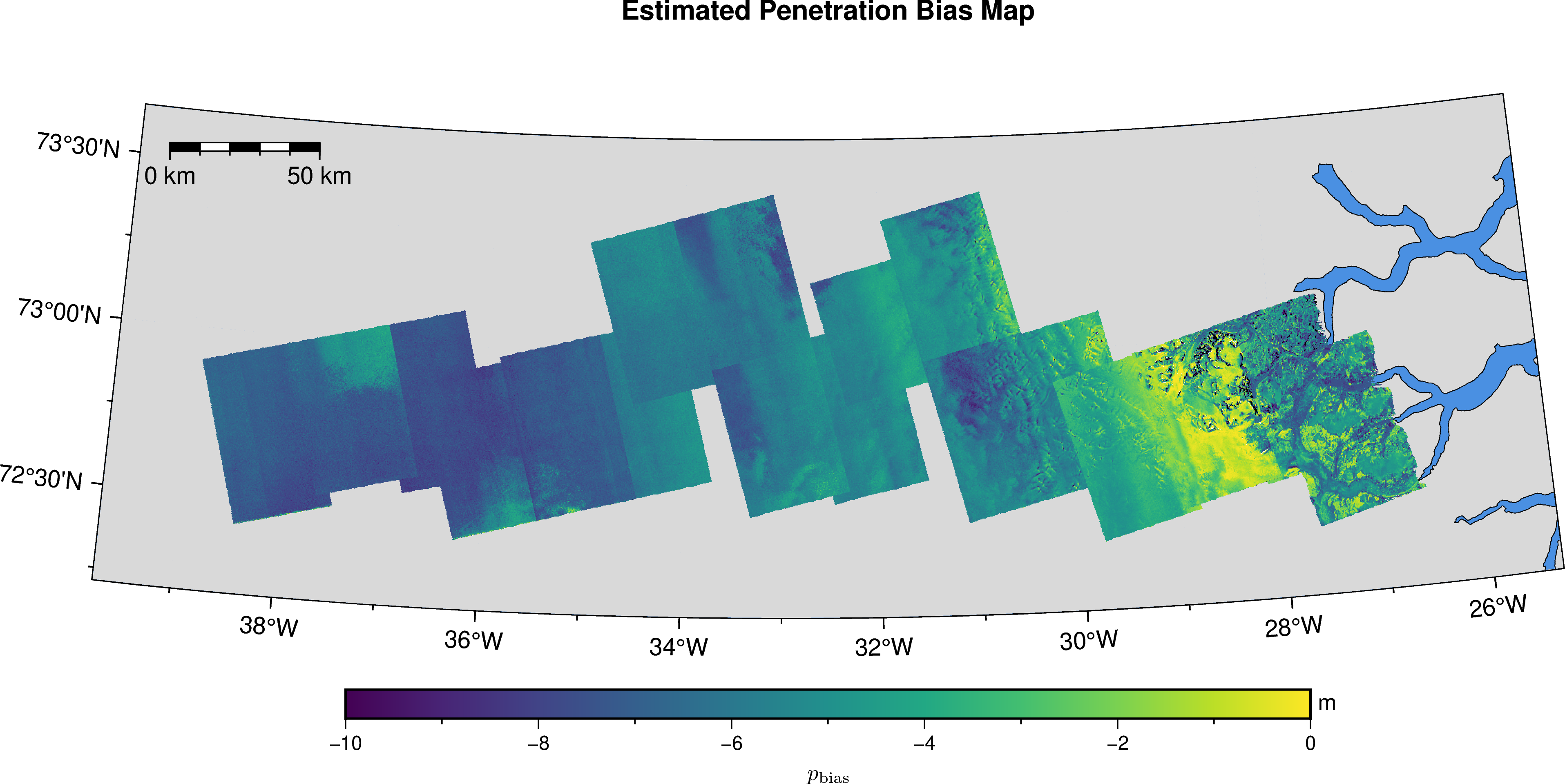}
        \end{minipage}
        \begin{minipage}{0.32\textwidth}
            \centering
            \includegraphics[width=\linewidth]{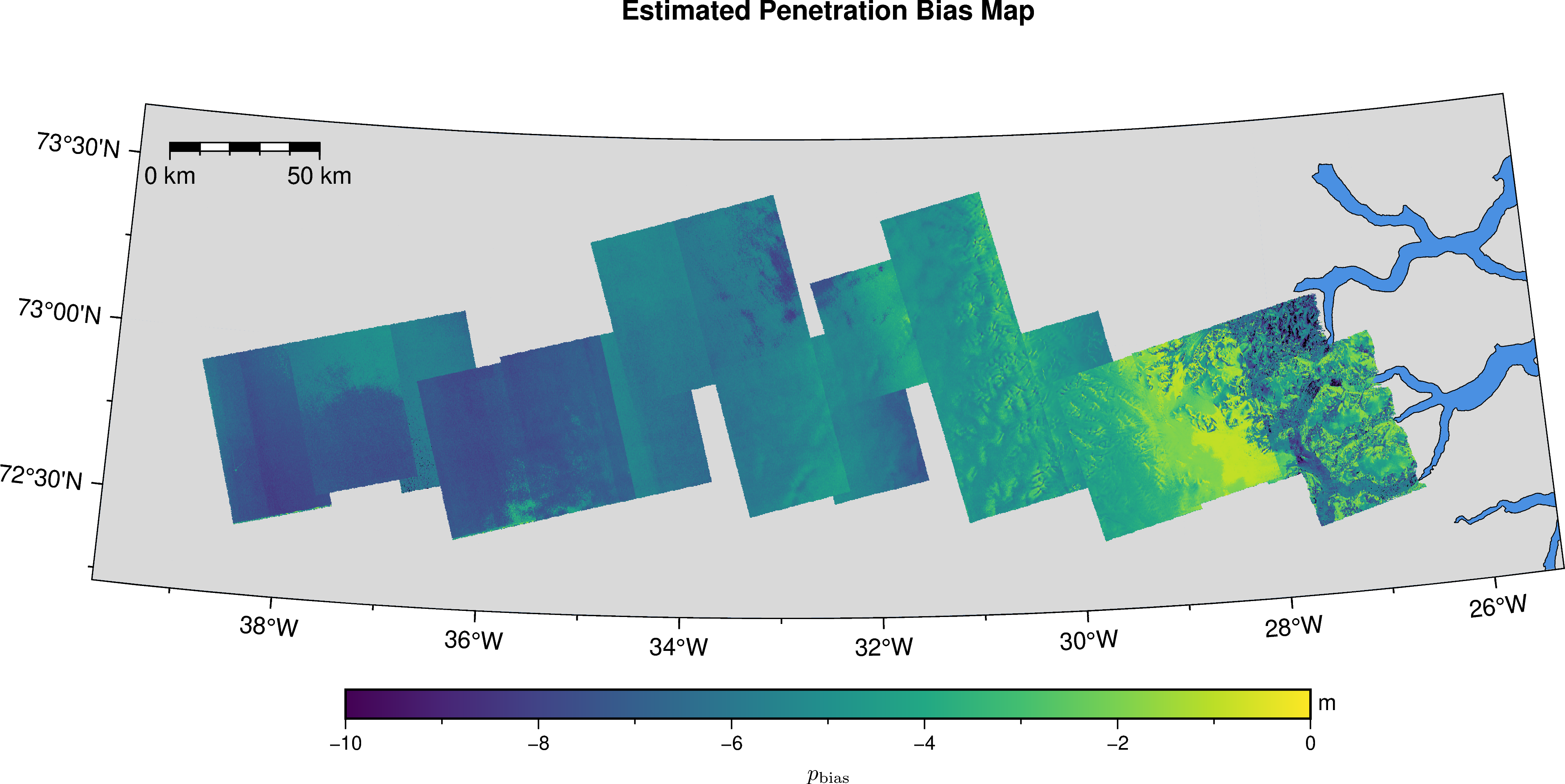}
        \end{minipage}
        \begin{minipage}{0.32\textwidth}
            \centering
            \includegraphics[width=\linewidth]{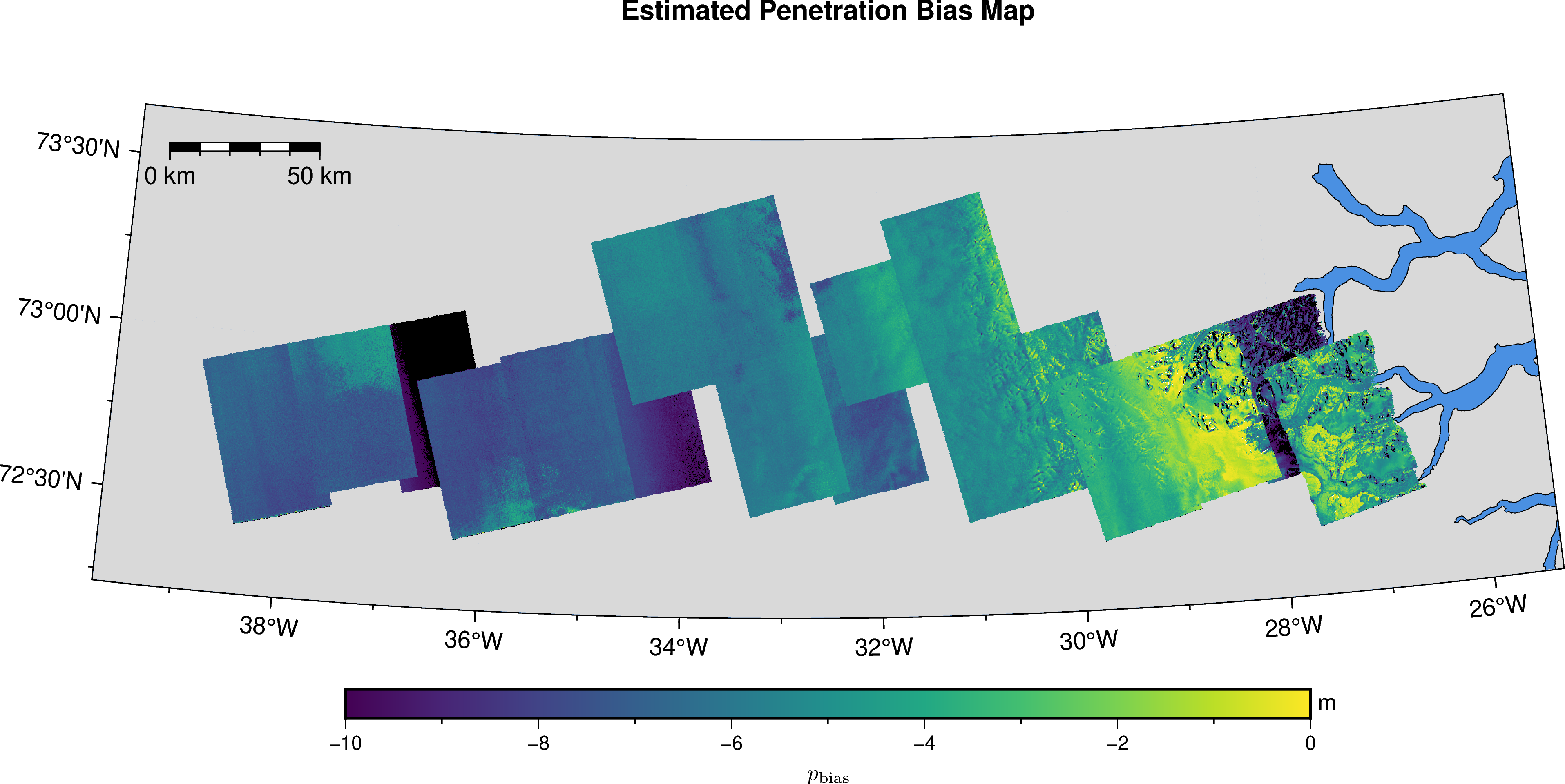}
        \end{minipage}
    
        \begin{minipage}{0.32\textwidth}
            \centering
            \includegraphics[width=\linewidth]{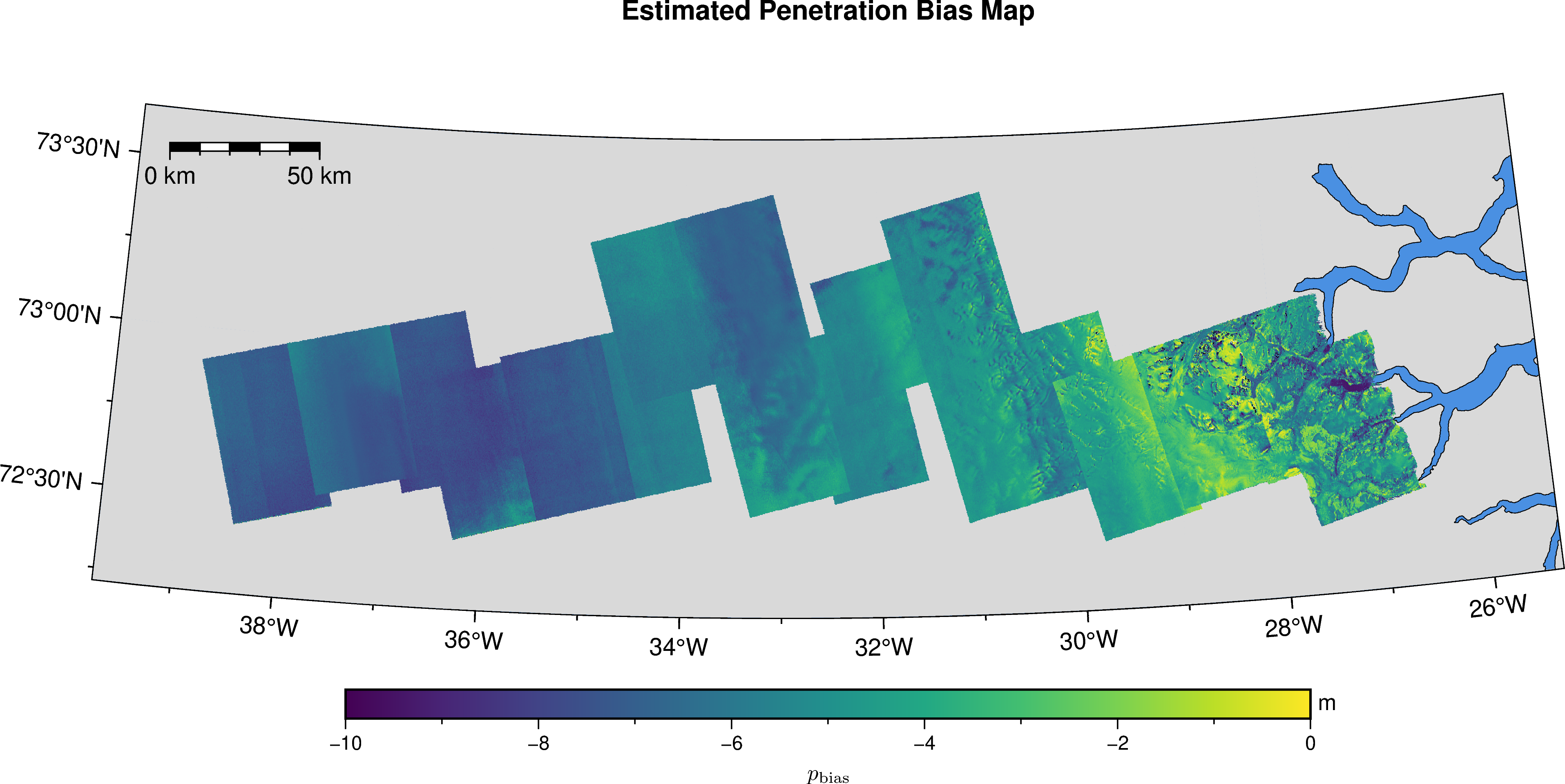}
        \end{minipage}
        \begin{minipage}{0.32\textwidth}
            \centering
            \includegraphics[width=\linewidth]{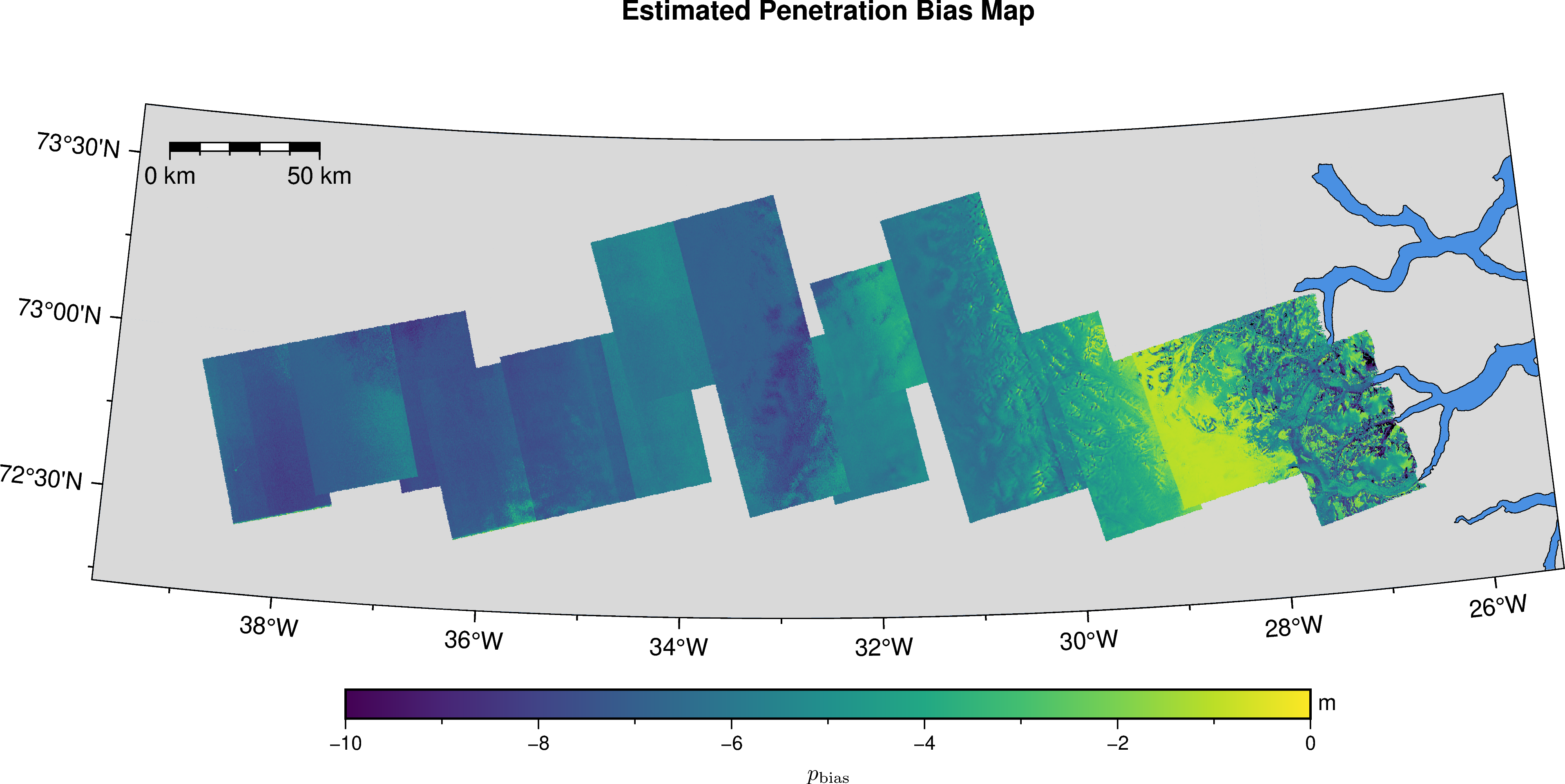}
        \end{minipage}
        \begin{minipage}{0.32\textwidth}
            \centering
            \includegraphics[width=\linewidth]{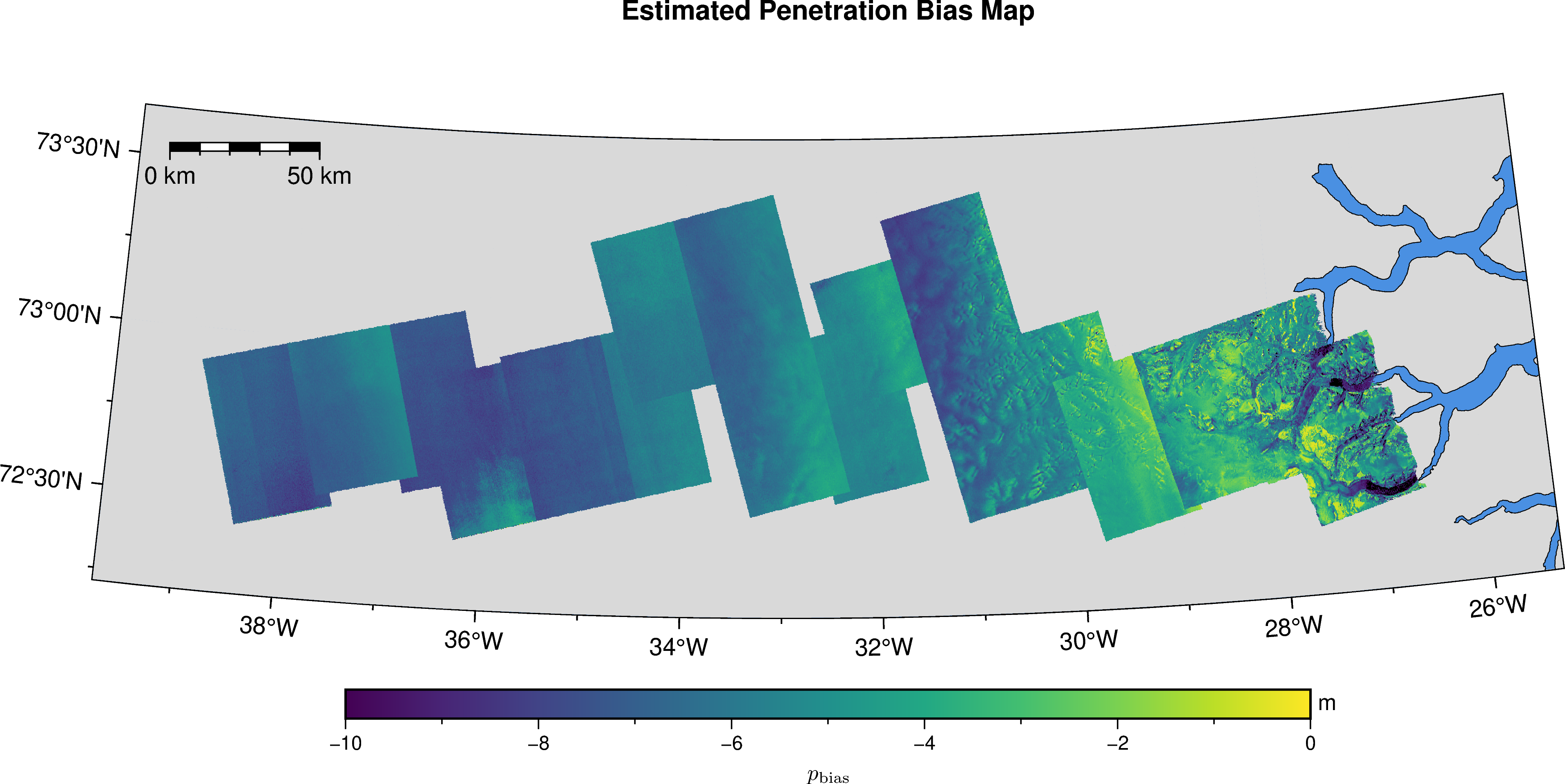}
        \end{minipage}
    
        \caption{Estimated penetration bias maps for the study region under three HoA training scenarios (rows) and three modeling approaches (columns). 
        \emph{Rows} (top to bottom): \emph{All}, \emph{Interpolation}, and \emph{Extrapolation} scenarios. 
        \emph{Columns} (left to right): \emph{Exponential}, \emph{Weibull}, and \emph{MLP} models. 
        Each panel shows the spatial distribution of the predicted bias (in meters), with blueish colors indicating deeper penetration bias.}
        \label{fig:map_penetration_models}
\end{figure*}

\begin{figure*}[t]
    \centering
    \begin{minipage}{0.30\textwidth}
        \centering
        \includegraphics[width=\linewidth]{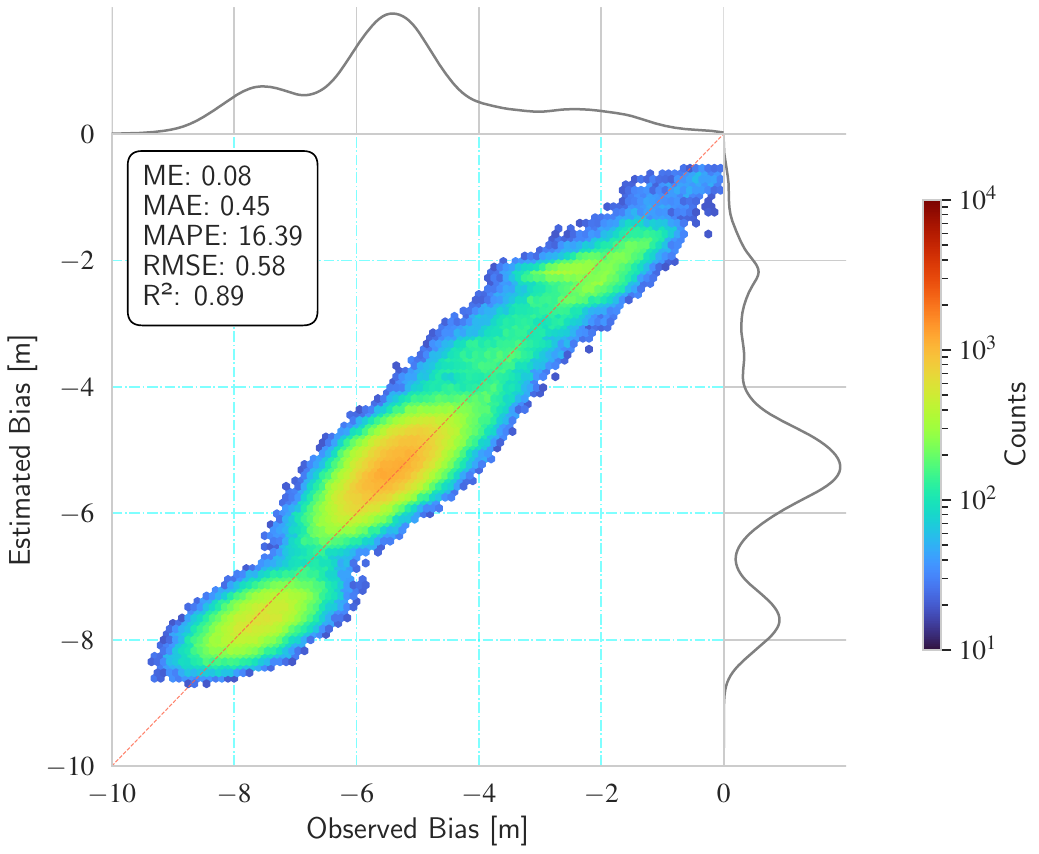}
    \end{minipage}
    \begin{minipage}{0.30\textwidth}
        \centering
        \includegraphics[width=\linewidth]{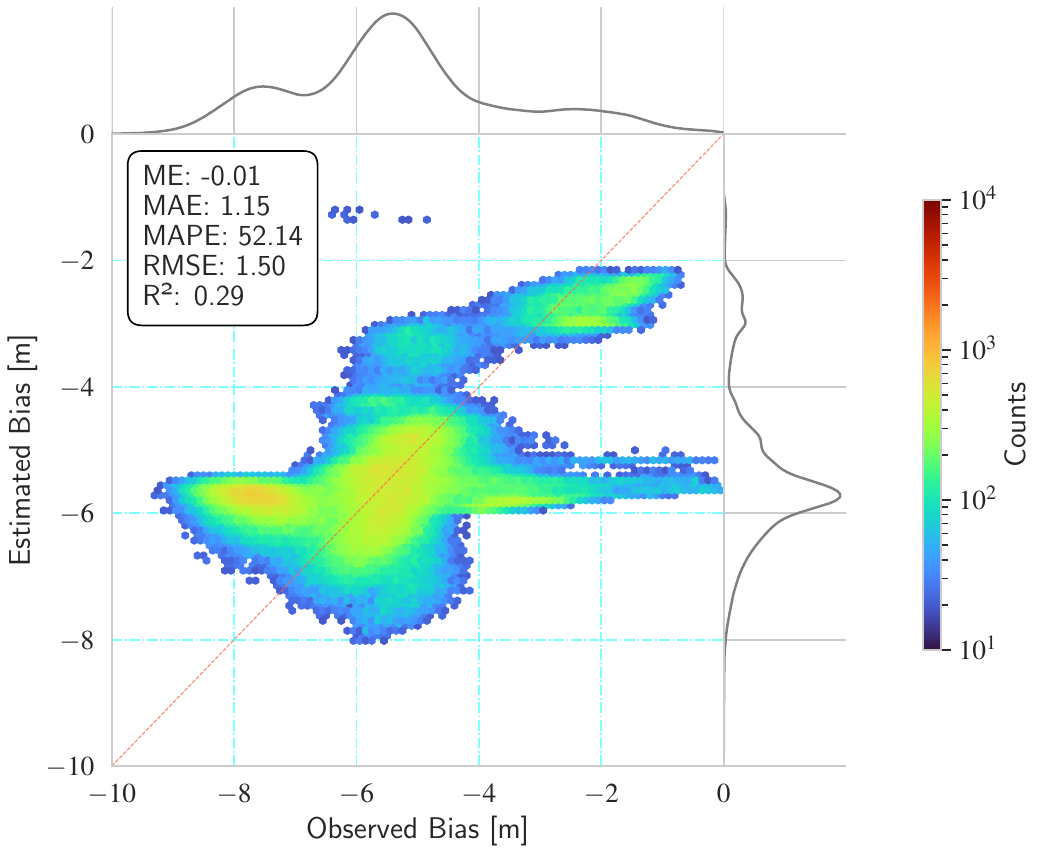}
    \end{minipage}
    \begin{minipage}{0.30\textwidth}
        \centering
        \includegraphics[width=\linewidth]{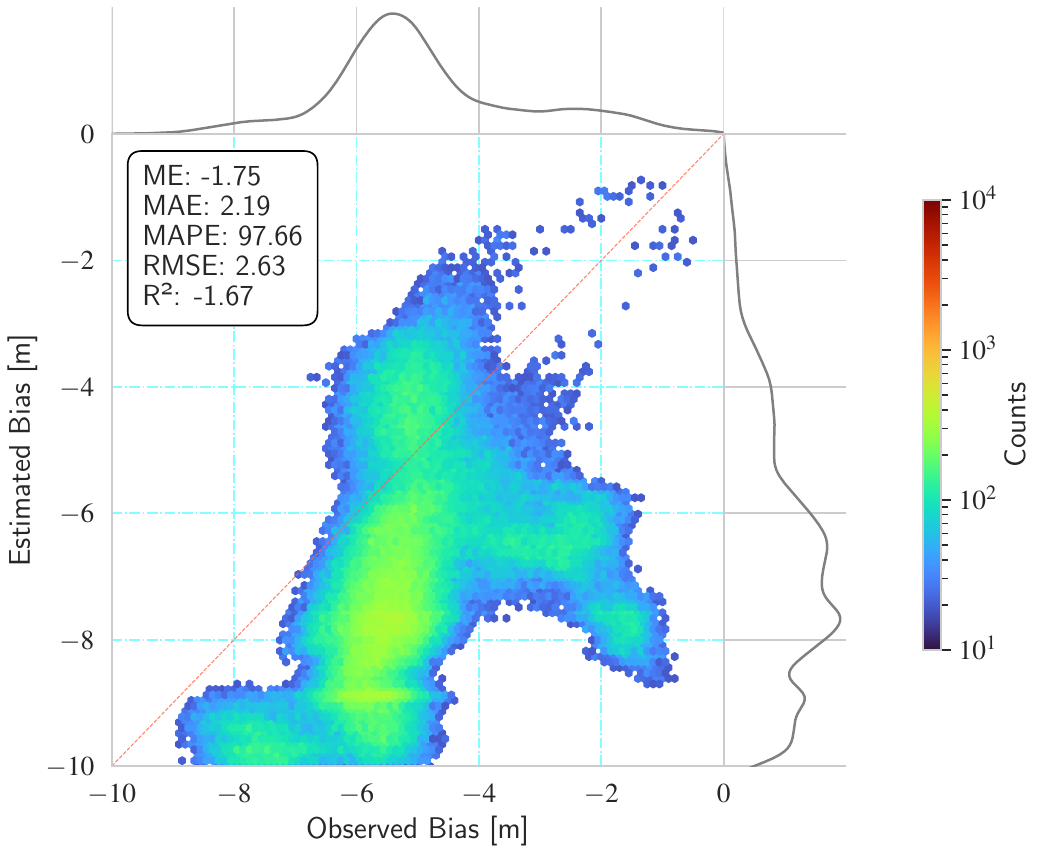}
    \end{minipage}

    \begin{minipage}{0.30\textwidth}
        \centering
        \includegraphics[width=\linewidth]{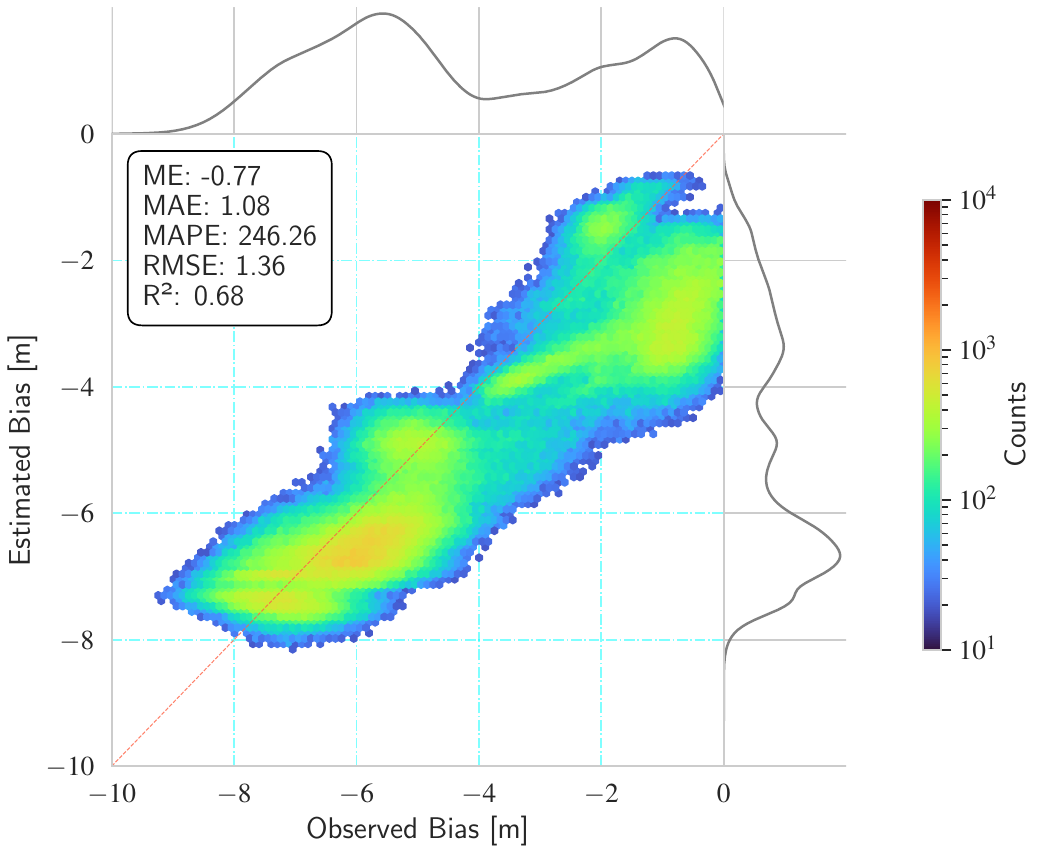}
    \end{minipage}
    \begin{minipage}{0.30\textwidth}
        \centering
        \includegraphics[width=\linewidth]{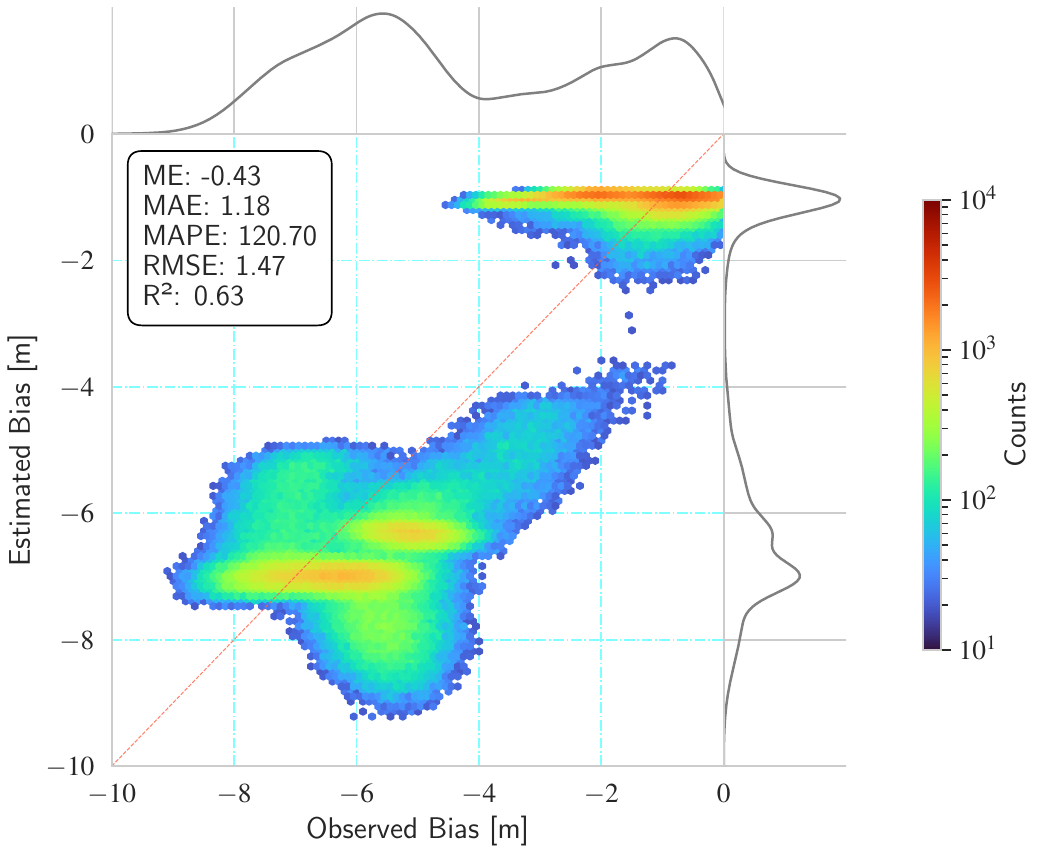}
    \end{minipage}
    \begin{minipage}{0.30\textwidth}
        \centering
        \includegraphics[width=\linewidth]{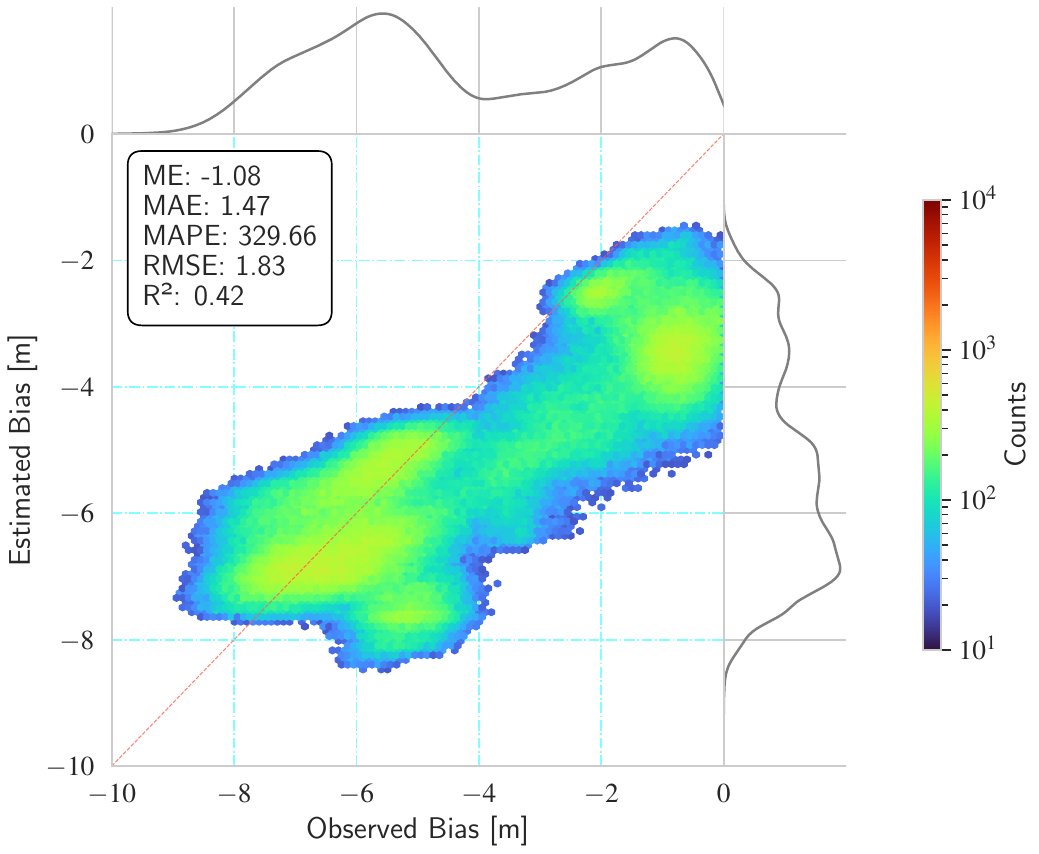}
    \end{minipage}

    \caption{Comparison of model estimations under different evaluation conditions showing the results only over the excluded scenes during training (unseen HoA scenes). 
    The \textbf{columns} represent different modeling approaches: 
    (Left) Hybrid Model with an Exponential Profile, 
    (Middle) Hybrid Model with a Weibull Profile, 
    (Right) Pure Machine Learning (ML) model using MLP. 
    The \textbf{rows} indicate different training scenarios: 
    (Left) Hybrid Model with an Exponential Profile, 
    (Middle) Hybrid Model with a Weibull Profile, 
    (Right) Pure Machine Learning (ML) model using MLP. 
    The \textbf{rows} indicate different training scenarios: 
    (Top) \emph{Interpolation} experiment; 
    (Bottom) \emph{Extrapolation} experiment.}
    
    \label{fig:2D_hist_models_no_training}
\end{figure*}

\begin{figure*}[!htb]
        \centering
        \begin{minipage}{0.32\textwidth}
            \centering 
            \includegraphics[width=\linewidth]{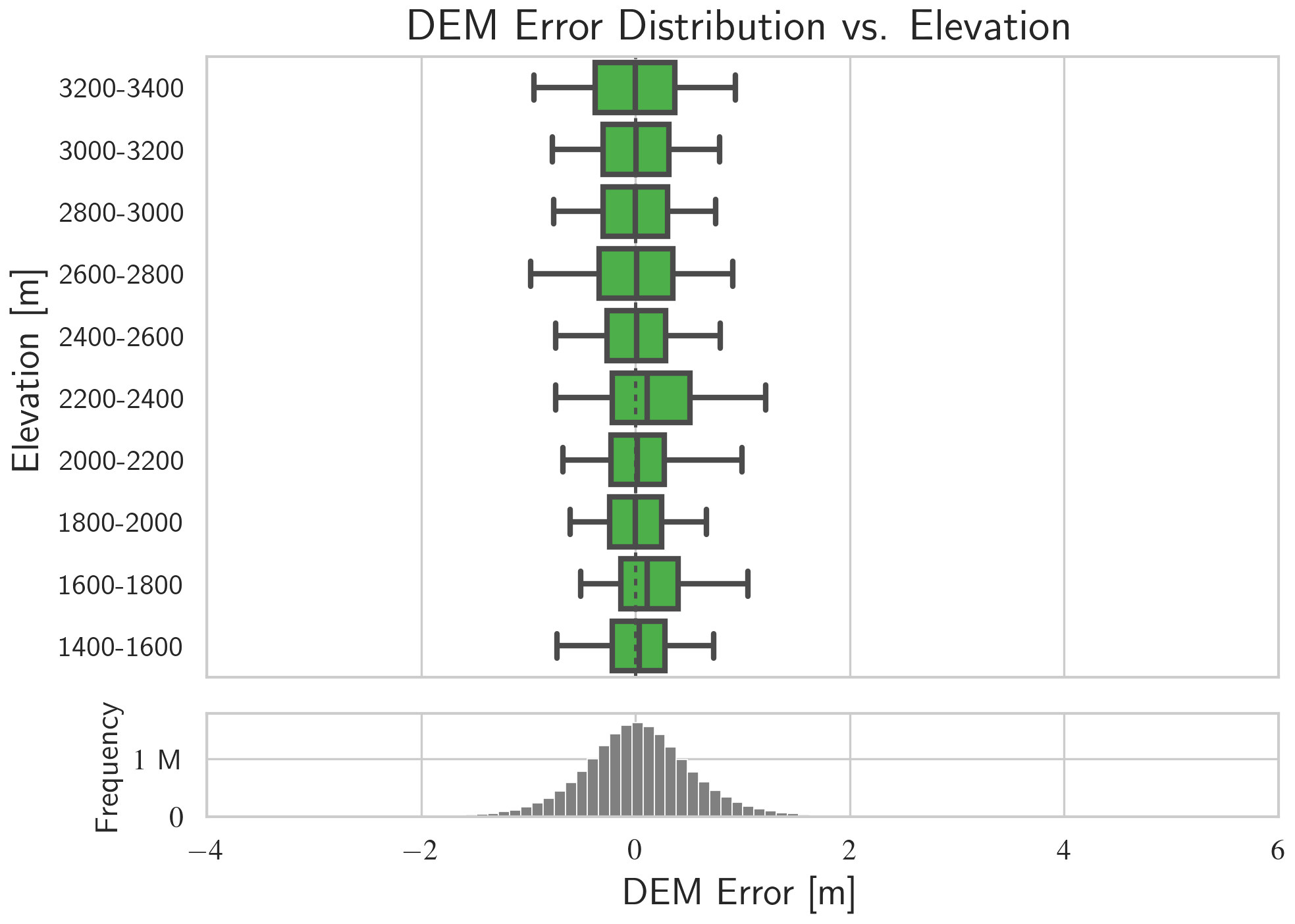}
        \end{minipage}
        \begin{minipage}{0.32\textwidth}
            \centering
            \includegraphics[width=\linewidth]{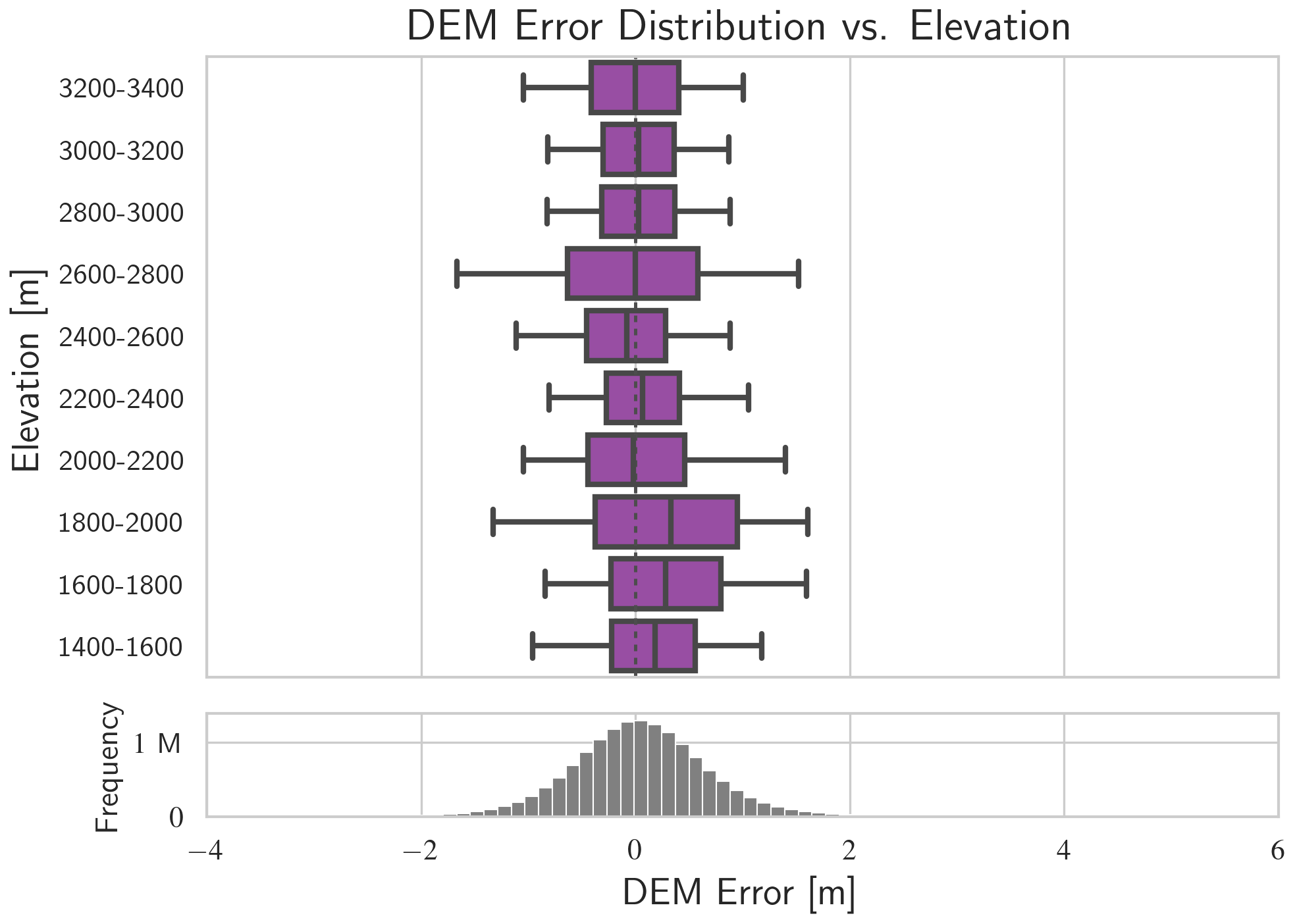}
        \end{minipage}
        \begin{minipage}{0.32\textwidth}
            \centering
            \includegraphics[width=\linewidth]{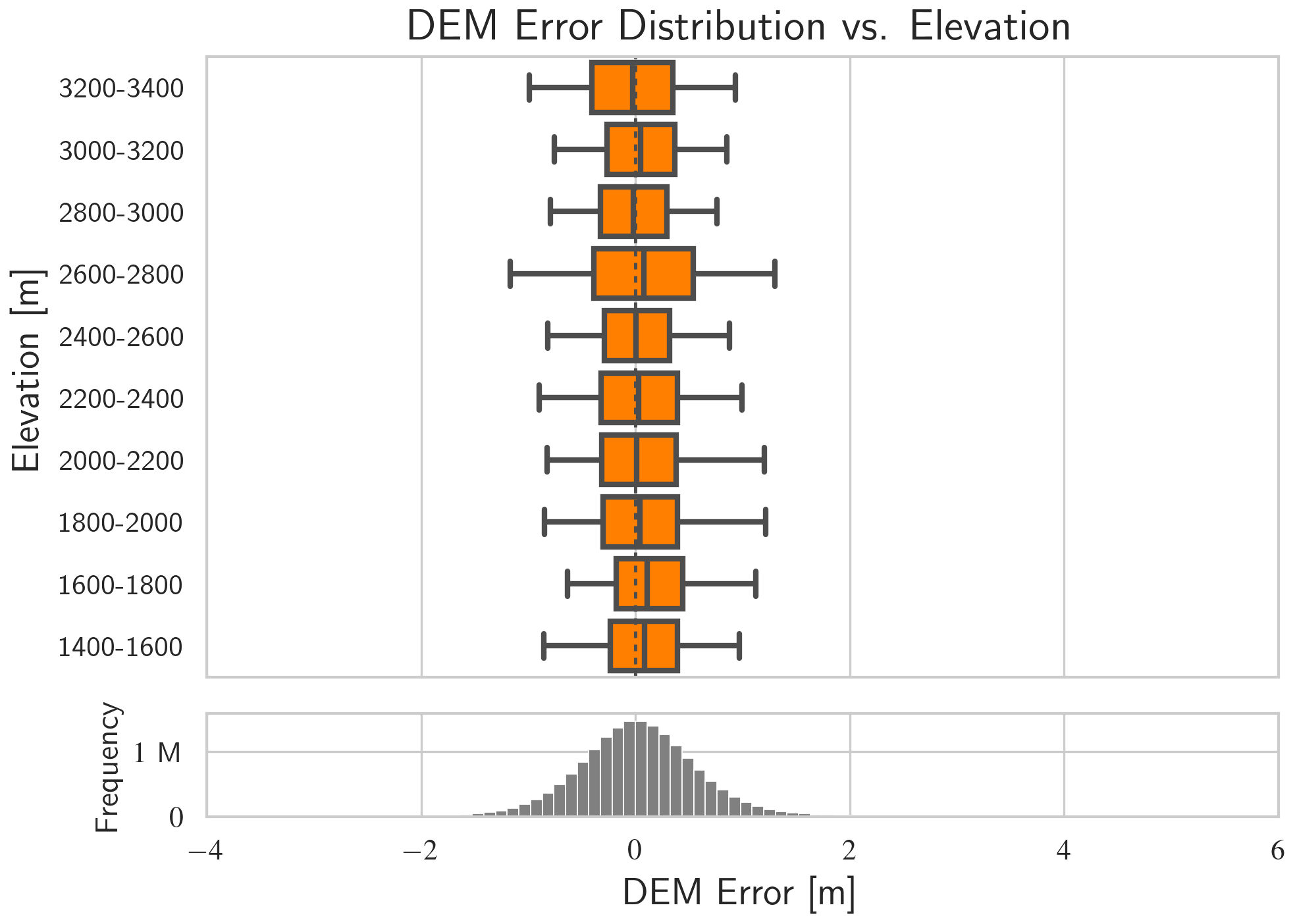}
        \end{minipage}
    
        \begin{minipage}{0.32\textwidth}
            \centering
            \includegraphics[width=\linewidth]{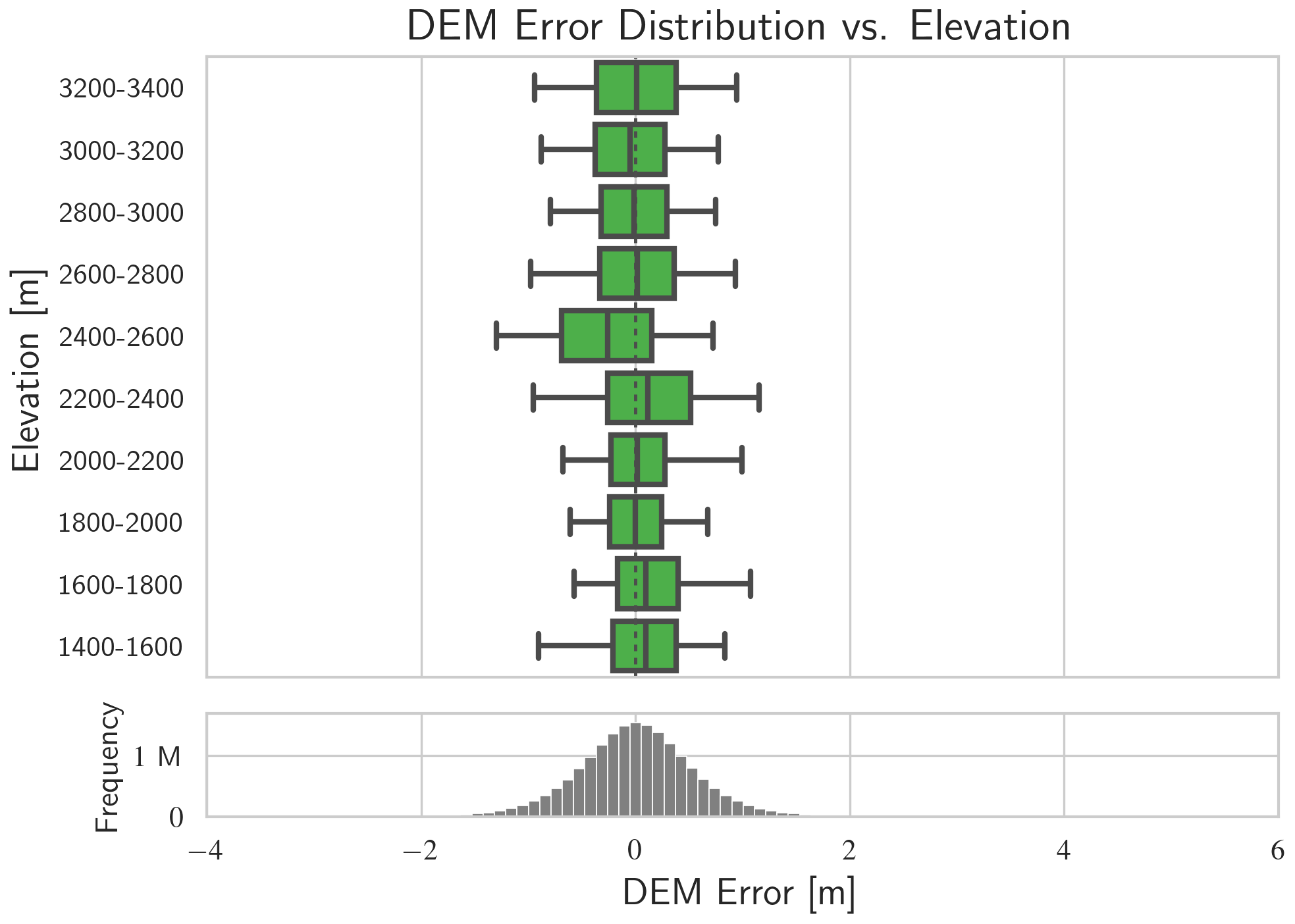}
        \end{minipage}
        \begin{minipage}{0.32\textwidth}
            \centering
            \includegraphics[width=\linewidth]{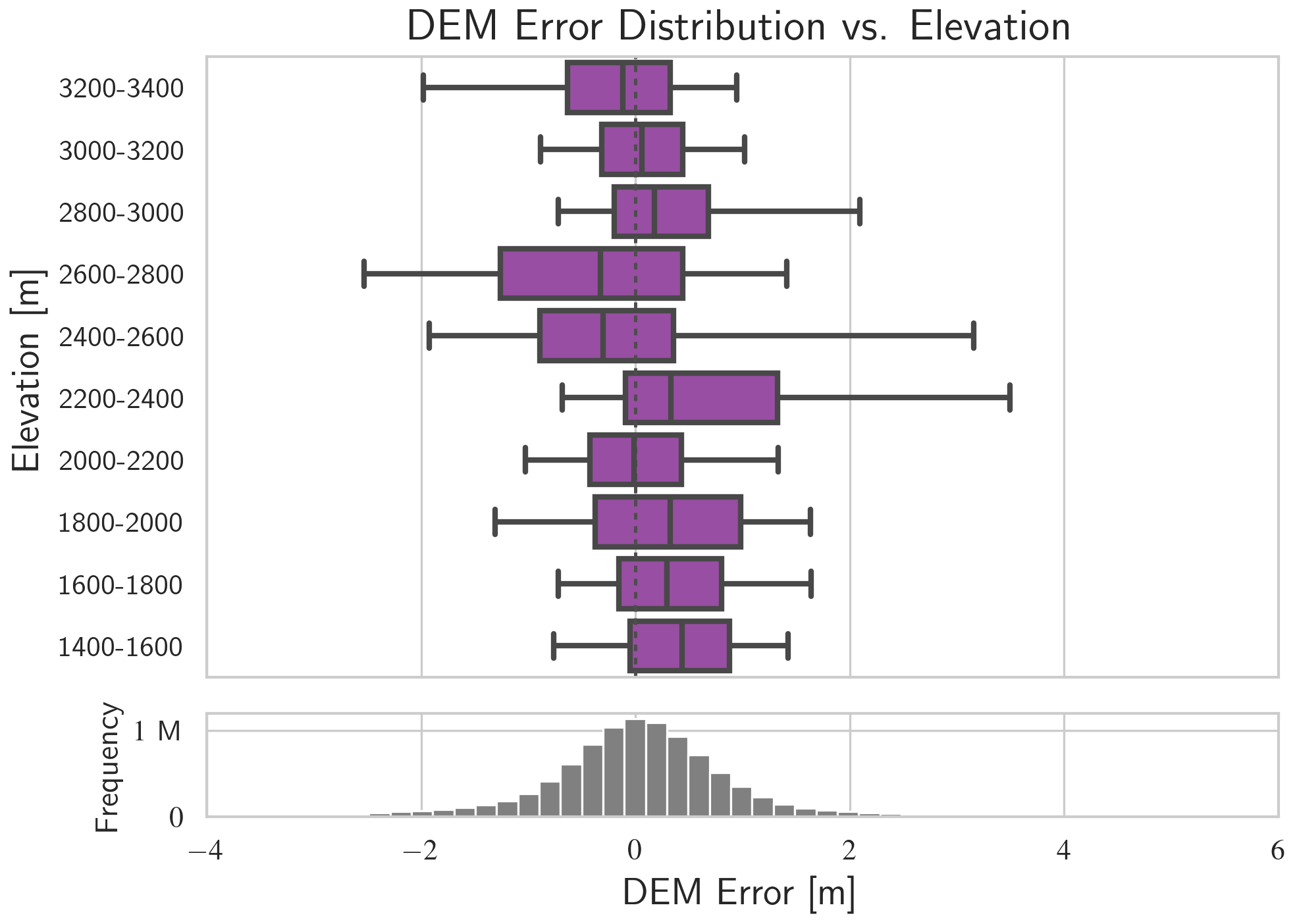}
        \end{minipage}
        \begin{minipage}{0.32\textwidth}
            \centering
            \includegraphics[width=\linewidth]{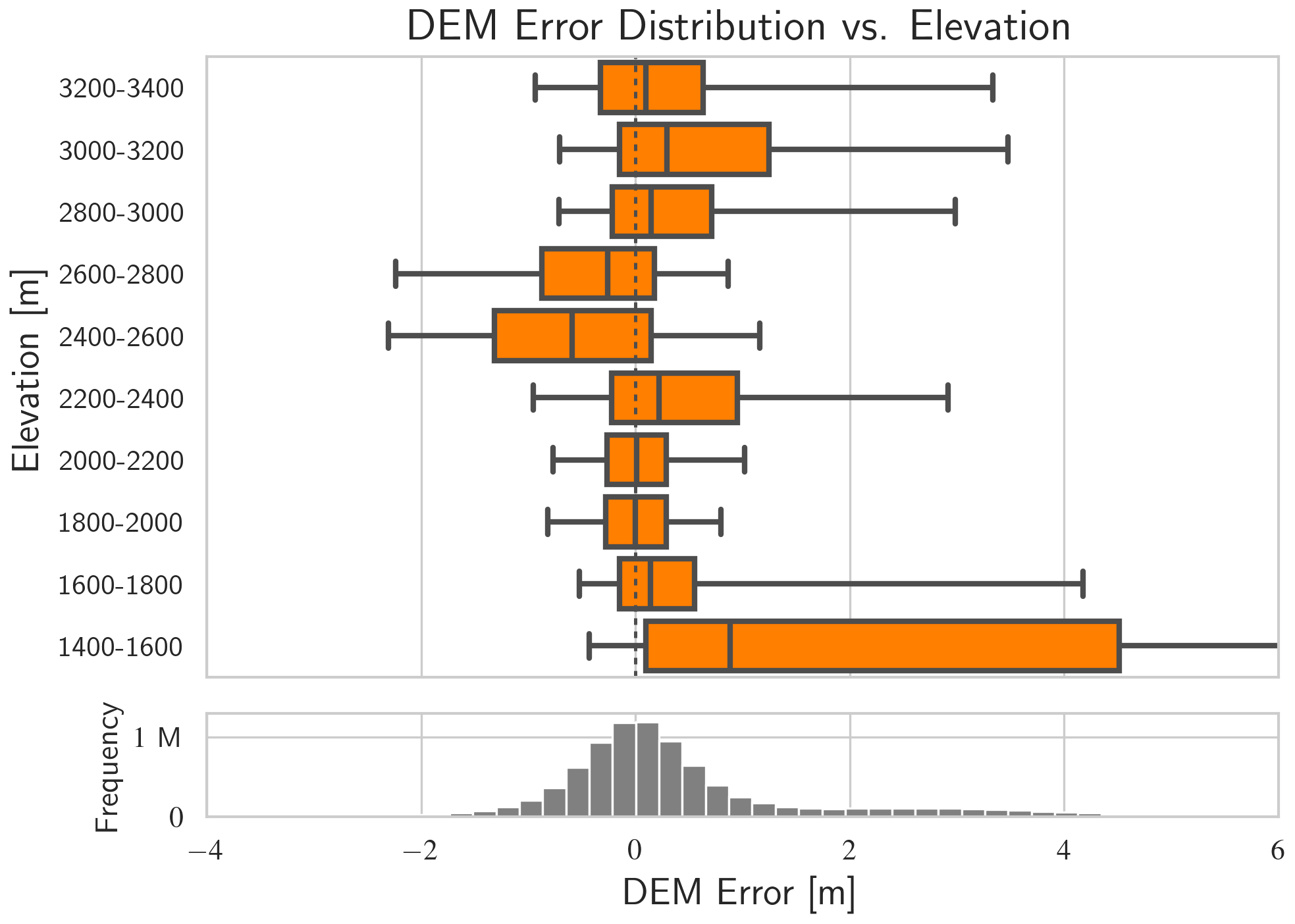}
        \end{minipage}
    
        \begin{minipage}{0.32\textwidth}
            \centering
            \includegraphics[width=\linewidth]{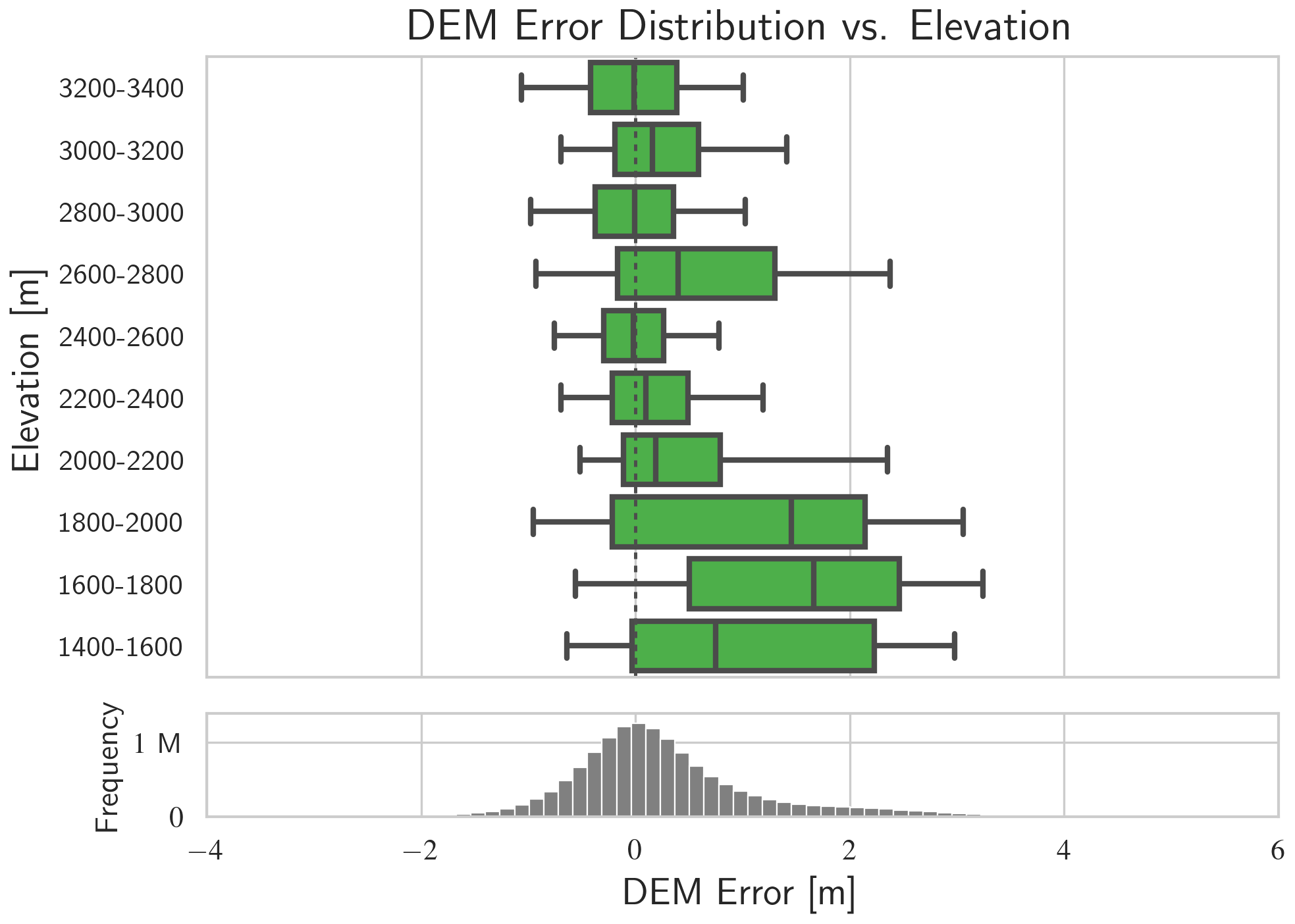}
        \end{minipage}
        \begin{minipage}{0.32\textwidth}
            \centering
            \includegraphics[width=\linewidth]{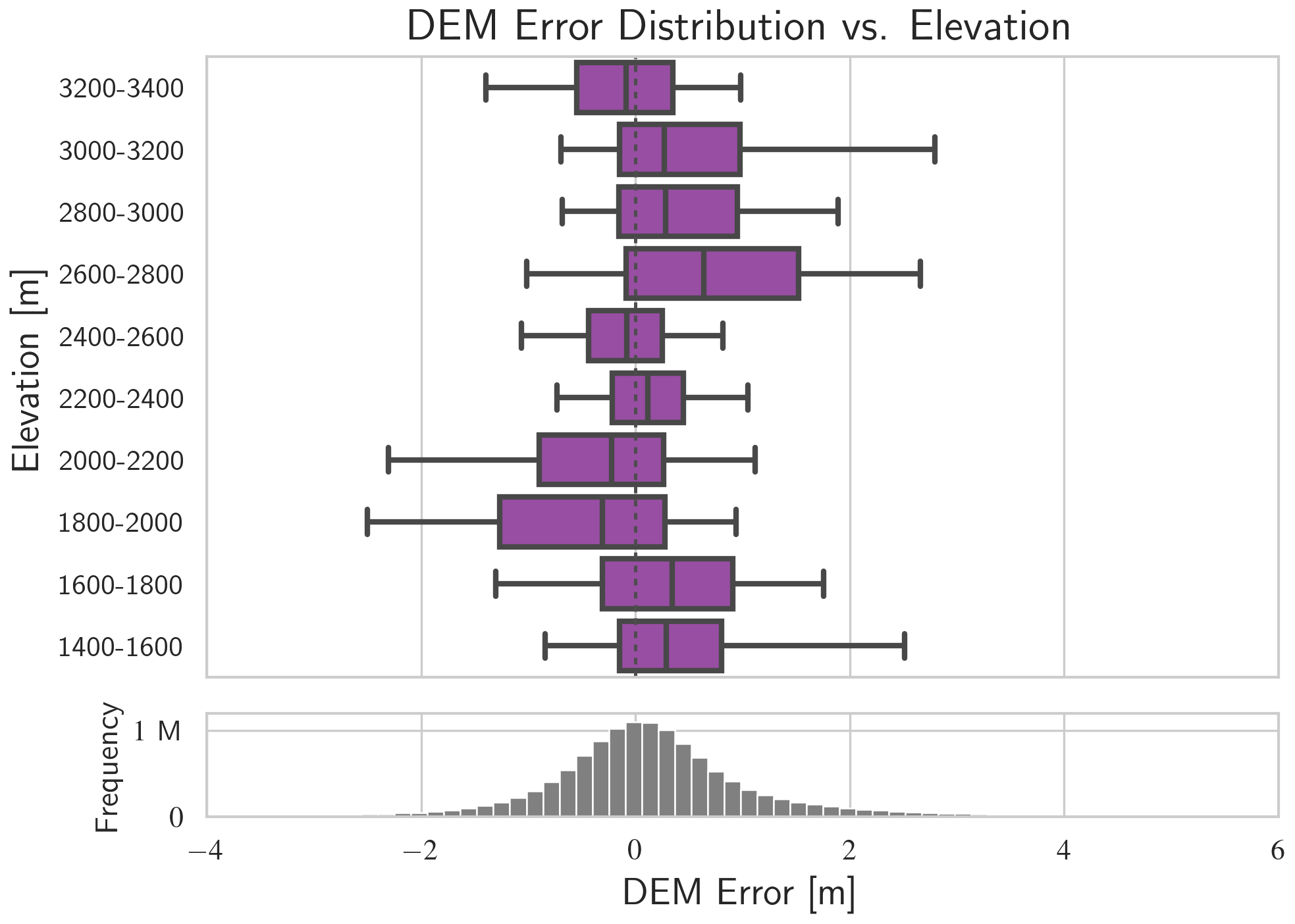}
        \end{minipage}
        \begin{minipage}{0.32\textwidth}
            \centering
            \includegraphics[width=\linewidth]{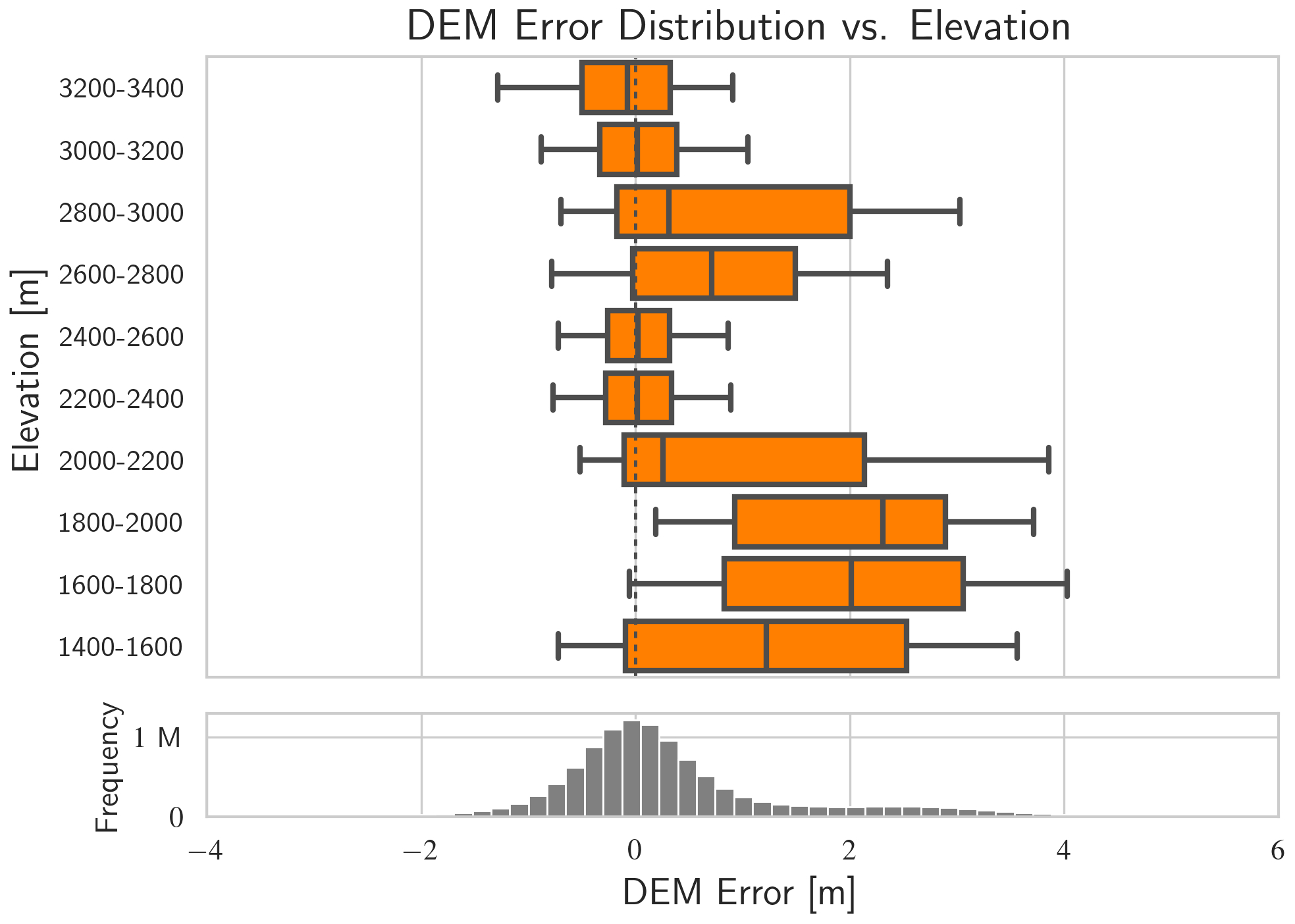}
        \end{minipage}
    
        \caption{Errors of the corrected DEMs, by compensating the original InSAR DEMs with the estimated penetration bias using different models and training scenarios. Shown as error distribution across elevation bins, computed using ATM LiDAR as reference under three HoA training scenarios (rows) and three modeling approaches (columns). 
        \emph{Rows} (top to bottom): \emph{All}, \emph{Interpolation}, and \emph{Extrapolation} scenarios. 
        \emph{Columns} (left to right): \emph{Exponential}, \emph{Weibull}, and \emph{MLP} models.
        }
        \label{fig:dem_elevation_error_histogram_all}
\end{figure*}

\end{document}